 \documentclass[nothing,twocolumn,10pt]{jmlr} % W&CP article 

\usepackage{booktabs}
\usepackage{siunitx}

\usepackage{enumitem}
\usepackage{multirow}
\usepackage{colortbl}

\definecolor{var1light}{HTML}{B3E0FF}
\definecolor{var1dark}{HTML}{A3C7D9} % Adjusted darkness
\definecolor{var2light}{HTML}{C7E9C0}
\definecolor{var2dark}{HTML}{BFD2A0} % Adjusted darkness
\definecolor{var3light}{HTML}{FFC0CB}
\definecolor{var3dark}{HTML}{FFB1B1} % Adjusted darkness
\definecolor{var4}{HTML}{E6D3FF}

\newcommand{\equal}[1]{{\hypersetup{linkcolor=black}\thanks{#1}}}

\theorembodyfont{\upshape}
\theoremheaderfont{\scshape}
\theorempostheader{:}
\theoremsep{\newline}

\jmlrpages{}

 \title[Unmasking the Chameleons: A Benchmark for OOD Detection in Medical Tabular Data]{Unmasking the Chameleons: A Benchmark for Out-of-Distribution Detection in Medical Tabular Data}

\author{
\Name{Mohammad Azizmalayeri} $^1$ \Email{m.azizmalayeri@amsterdamumc.nl}\\
\Name{Ameen Abu-Hanna} $^1$
\Email{a.abu-hanna@amsterdamumc.nl}\\
\Name{Giovanni Cin{\'{a}}} $^{1, 2, 3}$ 
\Email{g.cina@amsterdamumc.nl}\vspace{5pt}\\
\addr $^1$Department of Medical Informatics, Amsterdam Public Health Research Institute, Amsterdam UMC, University of Amsterdam, Netherlands. \\
\addr $^2$Institute of Logic, Language and Computation, University of Amsterdam, Netherlands. \\
\addr $^3$Pacmed, Amsterdam, Netherlands.
}

\begin{document}

\maketitle

\begin{abstract}
% Machine learning (ML) models have made remarkable advancements in analyzing medical data. 
% Nevertheless, their ability to generalize effectively to data originating from distributions that differ from the training distribution is limited. 
Despite their success, Machine Learning (ML) models do not generalize effectively to data not originating from the training distribution. 
To reliably employ ML models in real-world healthcare systems and avoid inaccurate predictions on out-of-distribution (OOD) data, it is crucial to detect OOD samples.
% In this paper, we address OOD detection within the context of medical tabular data. 
Numerous OOD detection approaches have been suggested in other fields - especially in computer vision - but it remains unclear whether the challenge is resolved when dealing with medical tabular data.
% Despite the numerous OOD detection approaches suggested in various fields - especially in computer vision - the challenge remains unresolved when dealing with medical tabular data. 
% Consequently, there is a pressing need for a comprehensive evaluation, in order to contrast these approaches in the medical context and understand their merits and limitations. 
To answer this pressing need, we propose an extensive reproducible benchmark to compare different methods across a suite of tests including both near and far OODs. 
Our benchmark leverages the latest versions of eICU and MIMIC-IV, two public datasets encompassing tens of thousands of ICU patients in several hospitals.
%Our benchmark incorporates up-to-date public datasets, including eICU and MIMIC-IV. 
We consider a wide array of density-based methods and SOTA post-hoc detectors across diverse predictive architectures, including MLP, ResNet, and Transformer. Our findings show that i) the problem appears to be solved for far-OODs, but remains open for near-OODs; ii) post-hoc methods alone perform poorly, but improve substantially when coupled with distance-based mechanisms; iii) the transformer architecture is far less overconfident compared to MLP and ResNet.

\end{abstract}

\section{Introduction}
\label{sec:Introduction}

\begin{figure}[t]
    \centering
    %\vskip 15 pt
    \includegraphics[width=\columnwidth]{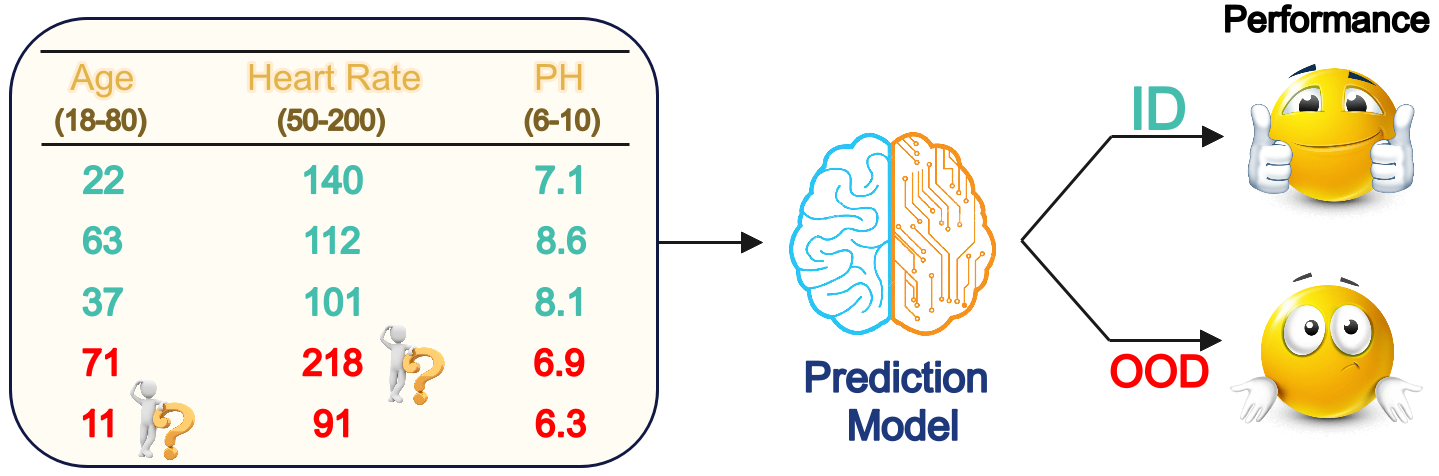}
    \caption{
    The practical dilemma of OOD data: there are no guarantees on how a model will perform on OOD data, hence real-time OOD detection becomes imperative.
    %Problem overview, where OODs cause an uncertain prediction.
    }
    \vskip -5 pt
    \label{fig:problem}
\end{figure}

The utilization of ML models in health-related applications is rapidly increasing \citep{greener2022guide, varoquaux2022machine}. However, a significant limitation lies in their performance evaluation, which is primarily based on optimizing the algorithms for data from the training distribution. This means that they may fail under distribution shift: a model trained on the data from a hospital may not generalize to other hospitals \citep{DKL_Miguel2021, de2023predicting}. 
Since such ML models are meant to be deployed in real-world healthcare scenarios, ensuring their reliability becomes of utmost importance. 
% As we aim to deploy these ML models in real-world healthcare scenarios, ensuring their reliability becomes of utmost importance. 
One way to prevent models from providing unreliable suggestions is to detect OOD samples in real time, prior to generating predictions. This is known as OOD detection, where a model trained on an in-distribution (ID) set is employed for distinguishing OOD samples from ID data. 
%Therefore, it is necessary to detect data coming from out of the training distribution prior to generating any prediction for it.

In this paper, we investigate this problem for tabular medical data. The problem has already been investigated mainly in the field of computer vision \citep{yang2021generalized, zimmerer2022mood}, but the results may not extend to tabular medical data. For example, while computer vision has focused on improving post-hoc OOD detectors \citep{yang2022openood}, it is demonstrated that these kinds of methods perform even worse than a random classifier in medical tabular data due to the phenomenon of over-confidence \citep{Ulmer2020Trust, ulmer2021know, zadorozhny2022out}. 

Existing OOD detection methods can be categorized into three main groups: i) post-hoc methods, which are detectors that can be applied on top of any trained classifier, such as Maximum Softmax Probability (MSP) \citep{hendrycks2017a}, ii) density-based methods, which are trained to estimate the marginal distribution of the training set in order to detect samples that fall out of the training distribution, such as auto-encoders (AEs) \citep{kingma2013auto}, and finally iii) methods that require retraining of the prediction model which are mainly designed specifically to be applied on images, such as OpenGAN \citep{kong2021opengan}. 

%The main goal in this work is to make a comprehensive study on these OOD detection methods within the medical tabular data and provide a fair comparison between them in this domain. 
The focus of this work is medical tabular data, thus we only consider the first two OOD categories since they can be applied to any medical tabular dataset without limitations. Furthermore, we examine three important architectures for the classifier on which post-hoc methods are implemented, namely MLP, ResNet, and FT-Transformer \citep{gorishniy2021revisiting}—to assess their impact on OOD detection.

A crucial aspect of comparing these methods involves evaluating their performance across diverse scenarios. Therefore, we incorporate the latest publicly available datasets including eICU and MIMIC-IV \citep{pollard2018eicu, johnson2023mimic} in our experiments. 
%Both datasets incorporate Intensive Care Unit (ICU) data, highlighting a shared aspect of this work.
%We also use different scenarios for splitting the datasets into the in/out set to account for the distance between the OOD sample and the training set. 
We also employ different approaches for classifying the data into ID and OOD sets. This lets us consider OOD sets near to and far from the IDs in the experiments, which are referred to as near-OOD and far-OOD \citep{fort2021exploring}. 
For example, to consider near-OOD instances, the ID and OOD sets can be constructed by dividing a dataset based on a distinguishing feature such as gender (male vs. female), and for a far-OOD experiment, we can use a dataset from a different data generating process, such as a distinct care protocol \citep{de2023predicting}, or generate artificial OOD samples.

%Finally, the methods are compared based on the separability of novelty scores they assign to ID and OOD samples.
We use this setup to conduct broad experiments that can illustrate the differences between various types of OOD detection methods in order to provide the community with an extensive reproducible benchmark in the medical tabular data domain\footnote{The codes are available at \url{https://github.com/mazizmalayeri/TabMedOOD}.}. Our results are provided in section \ref{sec:Results}, and the main findings are as follows: i) the OOD detection problem appears to be almost resolved on far-OOD samples, but it is still an open issue for near-OODs, ii) the distance-based post-hoc OOD detectors are better than other post-hoc methods, iii) unlike the claims in previous work, post-hoc methods can be competitive with the density-based ones, iv) the transformer architecture mitigates the problem of over-confidence that other models suffer from.

%ML in bio %Only good at training%reliable 
%OOD detection%we will check this in medical tab data (lack of a benchmark). Describe previous findings. %The CV field is going toward predictive while TabMed is not.
%Three kinds of OOD detectors
%We perform a benchmark using density and post-hoc. 3 artitechtures. 
%setup of oods.
%Main findings: 

\section{Related Work}
OOD detection has received a lot of attention, with several kinds of approaches being proposed for this purpose. This has resulted in interest in comparing these methods fairly within the same setting. Some benchmarks compare methods using standard image datasets \citep{han2022adbench, yang2022openood, zhang2023openood}. However, it is still necessary to have benchmarks that are specific to other domains and data modalities.
For example, the MOOD challenge \citep{zimmerer2022mood} investigated OOD detection within the context of medical imaging, with various new methods showing outstanding results \citep{marimont2021anomaly, tan2022detecting}. 

In the realm of medical tabular datasets, 
%which is our primary focus, 
notable endeavors have been undertaken. A pipeline is presented in \cite{nicora2022evaluating} to train a model on  ICU data and detect test samples for which the model might exhibit poor performance. BEDS-Bench \citep {avati2021bedsbench} has provided a benchmark on the generalization ability of ML models over electronic health records, highlighting performance drops under distribution shift. Moreover, it has been found that access to OOD data does not improve the test performance on  ICU data \citep{spathis2022looking}. The work by \cite{Ulmer2020Trust}, proposed one of the first comparisons between basic OOD detectors in the space of medical tabular data, pointing to the fact that the problem of OOD detection was wide open. Also, some guidelines have been provided on how to evaluate an OOD detector in practice within the context of medical data \citep{zadorozhny2022out} such as methods for selecting the OOD set during evaluation.
%and considerations in the deployment of an OOD detector.
%the interpretability of OOD detectors, how to select the OOD set for evaluation, and considerations in the deployment of an OOD detector. 
Compared to this related work, we benchmark the most recent SOTA OOD detection approaches and SOTA architectures, leading to novel insight e.g. on the over-confidence of transformers and the combination of post-hoc and distance-based methods.

% In contrast to this related work, we will benchmark the most recent SOTA OOD detection approaches and newer SOTA architectures like transformers. It is imperative to compare these approaches in-depth since they have advanced significantly over the last few years. Additionally, we analyze them within two up-to-date public medical tabular datasets.

\section{Problem Definition and Evaluation Protocol}
%\section{Problem Setup}
\label{sec:Problem}
%Assuming that the train and test samples in a dataset come from the same distribution, the main challenge is to detect any sample outside the training distribution. 
In the following, we describe the problem definition, the metrics used to measure performance, and how we select the ID and OOD sets.

\subsection{OOD Detection}
In training any ML model, we require a training set $\mathcal{D}_{in}=\{(x_i, y_i)\}_{i=1}^n$, where each instance $x_i$ has a label $y_i$. Our goal is to have a model $f:\mathcal{X}\rightarrow\mathcal{R}$ and a binary classifier $G_\lambda: \mathcal{X}\rightarrow\{0, 1\}$ such that for a test input $x\sim\mathcal{X}$:
%\vspace{-7pt}
\begin{equation}
    %\vspace{-5pt}
    G_\lambda(x; f) = \begin{cases}
      \text{OOD} \quad\quad f(x)\geq\lambda\\%[-0.8ex]
      \text{ID} \quad\quad f(x)<\lambda
    \end{cases}.
\end{equation}
%where samples that $f$ assigns a higher score are classified as the in distribution, and $\lambda$ is the threshold. 
The score according to $f$ is sometimes called the `novelty score'. Samples whose novelty score is higher than the threshold $\lambda$ are classified as OOD.
Hence, the final goal would be to train model $f$ such that it assigns lower scores to $x\sim\mathcal{D}_{in}$ and higher scores to $x\sim\mathcal{D}_{out}$, where $\mathcal{D}_{out}$ is the dataset that includes the OOD samples. 
%Clearly, the difficulty of this problem depends on the separability of $\mathcal{D}_{in}$ and $\mathcal{D}_{out}$.

\subsection{Metrics}
To assess the separability of the ID and OOD scores given by $f$, we measure the area under the receiver operating characteristic (AUROC) as a well-known threshold-independent classification criterion, and FPR@95, which measures the FPR at a $\lambda$ corresponding to the TPR being equal to 95\%.
%AUROC value is in the interval of $[0\%, 100\%]$, with a higher value indicating better separability.

\subsection{Near, Far, and Synthesized OOD} \label{sec:far_near_ood}
The distance between $\mathcal{D}_{in}$ and $\mathcal{D}_{out}$ plays an important role in the performance of OOD detection methods. 
It is relatively simpler to detect samples far from the training distribution than samples close to it; we refer to the former samples as far-OOD and to the latter as near-OOD.
To consider this in the comparisons, we define different sets of ID and OOD as follows.

\noindent\textbf{Near and far OODs:}
Assuming that we have a dataset $\mathcal{D}$, we can define the ID and OOD sets in two ways: 
%First, we can use a single feature such as gender (male vs. female), age (e.g, old vs young), ethnicity (e.g, Asian vs. American), and etc. to divide it into two parts that we can use as in and out sets.  
First, we can separate $\mathcal{D}$ based on a specific feature such as gender (male vs. female) or age (e.g. elderly vs. young). %ethnicity (e.g., Asian vs. American). 
This reflects what may happen when employing an ML model in practice. As an example, if one develops a model on a population of mostly young people, it might not be fully reliable when applied to the elderly.
The second way would be to use $\mathcal{D}$ as ID and a totally different dataset as OOD. Since identifying OODs in the second scenario appears to be easier, we will refer to the first scenario as near-OOD and the second as far-OOD following the convention in computer vision \citep{winkens2020contrastive, fort2021exploring}.

\noindent\textbf{Synthesized-OOD:} Following the data corruption suggested in \citet{Ulmer2020Trust}, we can simulate the OOD samples by scaling a single feature from ID set by a factor of 10, 100, or 1000. For each factor, we will repeat the experiments 100 times with different features, and average the results to isolate the influence of scaling in the results, minimizing the impact of the chosen feature. By increasing the scaling factor, it looks like we are gradually transforming near-OOD samples into far-OOD ones.

\section{Experiment Setup}
\label{sec:Experiment_Setup}
%In the following, the setting of experiments including methods and data is described.

\subsection{Supported OOD Detectors}\label{sec:Supported_OOD_Detectors}

Two main OOD detection method categories, described in Table \ref{tab:methods}, are included in our experiments. The first category is density estimators, which learn the marginal distribution of the ID data and label OOD samples by comparison to such distribution.
%and use that to label samples that fall out of the estimated density as OOD. 
%and use that as a measure of the novelty score for labeling the input.
We include 7 density-based models covering different types of density estimators. These methods are used in prior works and have reached outstanding results \citep{Ulmer2020Trust, zadorozhny2022out}. 

The second group is the post-hoc detectors, which can be integrated into any pre-trained classifier without requiring any additional fine-tuning or training. They mostly generate a novelty score for the input based on the classifier output or intermediate representations. We evaluate 17 different post-hoc detectors, including both commonly used detectors and top-performing ones in their respective fields.

\begin{table*}[t]
\small
\centering
\renewcommand{\arraystretch}{0.85}
\setlength{\tabcolsep}{2.5pt}
%\vskip -10pt
\begin{tabular}{ll}
\toprule
\textbf{Method}                    & \textbf{Short Description}                                                                 \\ \midrule
 AE, VAE \citep{kingma2013auto} &  
 Encodes input in a latent representation and reconstructs it.
 %Uses input reconstructing loss from the latent space as the novelty score.
 \\
 HI-VAE \citep{nazabal2020handling} &  Modifies VAE to consider heterogeneous data.\\
 Flow \citep{Papamakarios2017} \nocite{nflows} &   Estimates ID data by transformations into a normal distribution. \\
 PPCA \citep{tipping1999probabilistic} &  Reduces data dimensionality based on singular value decomposition.  \\
 LOF \citep{de2010finding} &  Compares local density of input to the density of its closest neighbors.   \\
 DUE \citep{van2021improving} &   Uses a deep neural Gaussian process for modeling the ID data.\\
 \midrule
MDS \citep{lee2018simple}                & Uses Mahalanobis distances to the class-conditional normal distributions.            \\
RMDS \citep{ren2021simple}               & Modifies MDS for detecting near-OODs.                                             \\
KNN \citep{sun2022out}                   & Measures distance of input to the $k_{\text{th}}$ nearest neighbor in the ID data. \\
VIM \citep{wang2022vim}                  & Uses logits and features norm simultaneously.                                     \\
SHE   \citep{zhang2023outofdistribution} & Measures the distance of input to the class-conditional representations.   \\
KLM   \citep{hendrycks2019anomalyseg}    & 
Uses KL distance of softmax output from its range over the ID data.
%Uses KL distance of softmax output with its respective ID data range.
\\
OpenMax   \citep{bendale2016towards}     & Fits a Weibull distribution on the logits instead of softmax.                     \\
MSP \citep{hendrycks2017a}               & Uses maximum softmax probability as a simple but effective baseline.                         \\
MLS   \citep{hendrycks2019anomalyseg}    & Uses the maximum logit score instead of MSP.                                      \\
TempScale   \citep{guo2017calibration}   & Calibrates the temperature parameter in the softmax.                              \\
ODIN   \citep{liang2018enhancing}        & Perturbs the input adversarially before using TempScaling.                        \\
EBO \citep{liu2020energy}                & Uses an energy function instead of softmax.                                       \\
GRAM   \citep{sastry2020detecting}       & Measures deviation of the Gram matrix from its range over the ID data.      \\
GradNorm   \citep{huang2021importance}   & Uses norm of the backpropagated gradients.                                             \\
ReAct \citep{sun2021react}               & Rectifies the model activations at an upper limit.                                \\
DICE \citep{sun2022dice}                 & Suggests sparsification of the last linear layer.                                 \\
ASH   \citep{djurisic2023extremely}      & Extends the DICE idea to the intermediate feature layers.                      \\ \bottomrule  
\end{tabular}
\caption{
%Post-hoc detectors evaluated in this work, with a short description.
Density-based models and post-hoc detectors evaluated in this work, with a short description.
}
\label{tab:methods}
\end{table*}

\subsection{Datasets} 
We use eICU \citep{pollard2018eicu} and MIMIC-IV \citep{johnson2023mimic} in our experiments as two public and popular datasets encompassing tens of thousands of ICU patients in several hospitals. Following the descriptions in section \ref{sec:far_near_ood}, each of these datasets is considered as far-OOD set for the other one. 
For the near-OOD case, we divide the datasets based on the time-independent variables, as this choice better emulates the potential distribution shifts in ID data that can be encountered in practice than the time-dependent ones. Among the time-independent variables available for each dataset, we have selected age (older than 70 as ID), gender (females as ID), and ethnicity (Caucasian or African American as ID) in eICU, and Age (older than 70 as ID), gender (females as ID), admission type (surgical in the same day of admission as ID), and first care unit (CVICU as ID) in MIMIC-IV dataset. In each case, the remaining part of the dataset would be OOD. The results for the age (older than 70), ethnicity (Caucasian), and first care unit (CVICU) are reported in the main text, and others in Appendix \ref{apd:near_ood}.

%For selecting the seperating feature for the near-OOD case, we used a Welch's t-test with $p<0.01$ to find the feature that causes more difference in the splitted parts, as otherwise it would be too difficult to identify OODs. In MIMIC-IV, we selected patients with first care unit equal to CVICU as the in-distribution and others as the OOD set since the t-test indicated $59.13\%$ difference between these sets. In eICU, we selected patients with ethnicity equal to Caucasian and others as OOD that indicated $56.08\%$ difference. Please note that the results for the Welch's t-test using other features are provided in the appendix and we have repeated the near-OOD case with some other features there.

\subsection{Pre-processing}

We pre-processed the eICU data using the pipeline provided in \citet{sheikhalishahi2020benchmarking}. Subsequently, the data is filtered to keep only patients with a length of stay of at least 48 hours and an age greater than 18 years old. Additionally, data with unknown discharge status were removed. Furthermore, patients with NaN values in the features used in our models are also removed. This pre-process resulted in a total $54826$ unique patients for this dataset.

For the MIMIC-IV, we used the pipeline provided in \citet{gupta2022extensive} with slight modifications e.g., we added a mapping from the feature IDs to feature names to have each feature name in the final pre-processed data. The data is then filtered similarly to the eICU dataset. This resulted in $18180$ unique patients for this dataset.

These datasets contain different types of clinical variables, but they are not recorded for all the patients. To avoid NaN values as much as possible, we are interested in using only the more important variables that are recorded for more patients. 
%In these datasets, there are different types of clinical variables that are not available for all patients. Hence, we are interested in using only the more frequent ones to avoid the existence of NaN values in the selected variables for a patient as much as possible. 
Based on these criteria and considering the important clinical variables suggested in \citet{Ulmer2020Trust}, we have selected a combination of time-dependent and time-independent variables for each dataset, which are reported in  Appendix \ref{apd:variables}. 
\iffalse
%For each variable, we have first stated its name in eICU and after that its name in MIMIC-IV. The variables that are only used in one of these datasets are shown with $\times$ for the other one.
%\begin{itemize}[itemsep=0pt,parsep=0pt, topsep=1pt, itemindent=-8pt]
%\end{itemize}
%\vskip 2pt 
%\noindent
%Time dependent: (pH, pH), (Temperature (C), Temperature), (Respiratory Rate, Respiratory rate), (O2 Saturation, Oxygen saturation), (MAP (mmHg), Mean blood pressure), (Heart Rate, Heart Rate), (glucose, Glucose), (GCS Total, $\times$), (Motor, $\times$), (Eyes, $\times$), (Verbal, $\times$), (FiO2, $\times$), (Invasive BP Diastolic, Diastolic blood pressure), (Invasive BP Systolic, Systolic blood pressure)
%\vskip 2pt 
%\noindent
%Time independent: (gender, gender), (age, age), (ethnicity, $\times$), (admissionheight, $\times$), (admissionweight, $\times$), ($\times$, admission\_type), ($\times$, first\_careunit)
%\vskip 2pt 
\fi

It should be noted that when datasets are evaluated against each other, only the variables found in both datasets are taken into account. Moreover, for the time-dependent variables, we aggregated the time series by means of 6 different statistics including mean, standard deviation, minimum, maximum, skewness, and number of observations calculated over windows consisting of the full time-series and its first and last 10\%, 25\%, and 50\%.

\subsection{Task and Prediction Models}

To perform post-hoc OOD detection, we would need a prediction model performing a supervised classification task.
The main task that is used in prior work is mortality prediction \citep{sheikhalishahi2020benchmarking, Ulmer2020Trust, meijerink2020uncertainty, zadorozhny2022out}. In mortality prediction, we only use the first 48 hours of data from the intensive care unit collected from patients to predict the in-hospital mortality. It is noteworthy that the mortality rate in the pre-processed data is $12.57\%$ in the MIMIC-IV and  $6.77\%$ in the eICU dataset.

To perform this task, we consider three widely used architectures: MLP, ResNet, and FT-Transformer \citep{gorishniy2021revisiting}. MLP passes data through the fully-connected layers and non-linear activation functions with dropout to improve the generalization. ResNet adds batchnorm and residual connections to MLP. 
%Hence, we can say that ResNet is the improved version of MLP in the classification task. 
We also consider FT-Transformer, constituted of transformer blocks that utilize the attention mechanism \citep{vaswani2017attention}. 
%Following the guidelines in the original paper, these models are trained for $10$ epochs with a batch size equal to 64 and AdamW \citep{loshchilov2018decoupled} optimizer with a learning rate equal to $1e\text{-}3$. 
%The mortality prediction performance of these models is reported in the Appendix \ref{apd:mortality}.

%Datasets and preprocess feature eng.
%Training Hyperparams
%Task and Predictive model artitechture
%Define far-ood and near-ood for each dataset (only the names. The methodology of defining ood is discussed in section 2.)

\section{Results}
\label{sec:Results}
%Far-ood
%Near-OOD
%Synthesied OOD
%Over-confidence
%Missclassified as OOD

In this section, we describe the results for each of the OOD settings based on the AUROC criterion. Results for FPR@95 are presented in Appendix \ref{apd:fpr}.  
Moreover, each experiment is repeated 5 times, and the results are averaged to reduce the impact of randomness in selecting train/test data and training itself. Additionally, we have measured the mortality prediction performance of the prediction models in Appendix \ref{apd:mortality}, indicating that they are trained well.

\subsection{Far-OOD}

Results for the far-OOD setting are displayed in Table \ref{tab:near_far_ood}.
%, where each result is averaged over 5 runs to reduce the impact of randomness in selecting train/test data and training itself. 
According to this table, there are methods that can effectively detect OOD data on each dataset. 

Among the density-based methods, Flow attains the best result on eICU, while others exhibit superior performance on MIMIC-IV. Additionally, DUE can detect OODs on MIMIC-IV, but it falls short of being competitive on the eICU dataset. Except for these two approaches and HI-VAE, other density-based methods including AE, VAE, PPCA, and LOF demonstrate strong performance on both datasets.

Within the post-hoc methodologies, MDS exhibits better results compared to others regardless of the choice of the prediction model. Moreover, MDS applied on ResNet is competitive with the density-based approaches, even marginally outperforming them based on the average results across both datasets. After MDS, there is not a single winner. However, approaches like KNN, VIM, and SHE which somehow compute the distance to the training set as the novelty score,  outperform the rest. 

Regarding the prediction model, the top-performing methods on ResNet demonstrate superior results compared to both MLP and FT-Transformer. However, a problem with ResNet and MLP is that they have over-confidence issues with some approaches on MIMIC-IV. This means that they have more confidence in the OODs, resulting in a performance even worse than a random detector. This is mainly being observed with the detectors that do not rely on distance-based novelty scores such as MSP and GradNorm. Note that the eICU dataset contains data from a more extensive array of hospitals, which can increase diversity in the data and reduce the over-confidence on this dataset.

FT-Transformer seems to solve the over-confidence problem observed in MLP and ResNet, as all detectors perform better than random on both datasets. The attention mechanism in the transformer blocks enables this model to consider relationships between input elements, leading to a better understanding of ID data.

\subsection{Near-OOD}

Results for the near-OOD setting are presented in Table \ref{tab:near_far_ood}. 
In the eICU dataset, the diversity among the ID data and the proximity of OODs to ID have collectively yielded an almost random performance for all the approaches. Still, it indicates that methods like MDS and Flow are marginally better than others.

In MIMIC-IV, which contains less diversity,  the age variable still results in an almost random performance across all approaches, but FCU reflects some differences between detectors. Among the density-based approaches, AE and VAE are the best choices, followed by PPAC, LOF, and HI-VAE. The post-hoc methods mostly demonstrate similar performance within the same architecture category. Moreover, they can be competitive with density-based methods when applied on the FT-Transformer.

\begin{table*}[]
\footnotesize
%\scriptsize
\centering
\renewcommand{\arraystretch}{0.85}
\setlength{\tabcolsep}{4pt}
\begin{tabular}{c|c|ccc|ccc}
\toprule
\multirow{2}{*}{Model}          & \multirow{2}{*}{Method}    & \multicolumn{3}{c|}{eICU}                      & \multicolumn{3}{c}{MIMIC-IV}                        \\
               &           & Far-OOD     & Near-OOD(Eth) & Near-OOD(Age) & Far-OOD     & Near-OOD(FCU) & Near-OOD(Age) \\ \midrule
%\multirow{6}{*}{-}              
\rowcolor{var4} & AE        & 96.5$\pm$0.2  & 55.1$\pm$0.8     & 50.0$\pm$0.4     & 99.8$\pm$0.0  & 79.4$\pm$0.6           & 56.8$\pm$0.4     \\
\rowcolor{var4}               & VAE       & 95.8$\pm$0.2  & 55.4$\pm$0.7     & 50.1$\pm$0.4     & \textbf{99.8$\pm$0.0}  & \textbf{79.7$\pm$0.6}          & \textbf{57.0$\pm$0.4}     \\
\rowcolor{var4}  &    HI-VAE   &  56.7$\pm$1.6 &   44.3$\pm$1.2  & 44.4$\pm$2.1     & 68.8$\pm$8.8 &     79.1$\pm$13.4       &    56.8$\pm$10.0  \\

\rowcolor{var4}  Density          & Flow      & \textbf{100.0$\pm$0.0} & \textbf{61.1$\pm$0.9}     & 49.7$\pm$0.4     & 87.4$\pm$7.5  & 31.8$\pm$3.9           & 51.6$\pm$0.5     \\
\rowcolor{var4}      Based         & PPCA      & 96.7$\pm$0.2  & 59.1$\pm$0.6     & \textbf{51.3$\pm$0.5}     & 99.8$\pm$0.0  & 75.1$\pm$0.6           & 56.8$\pm$0.7     \\
\rowcolor{var4}               & LOF       & 96.5$\pm$0.1  & 56.0$\pm$0.8     & 49.5$\pm$0.5     & 99.2$\pm$0.2  & 73.5$\pm$0.5           & 55.3$\pm$0.8     \\
\rowcolor{var4}               & DUE       & 73.4$\pm$0.5  & 53.7$\pm$0.9     & 49.5$\pm$0.4     & 98.2$\pm$0.2  & 59.8$\pm$1.7           & 51.6$\pm$0.6     \\
\midrule
\rowcolor{var1light} \multirow{17}{*}{MLP}          & MDS       & \textbf{84.3$\pm$1.4}  & \textbf{56.8$\pm$0.9}     & \textbf{51.7$\pm$0.7}     & \textbf{98.9$\pm$0.6}  & 68.0$\pm$3.9           & \textbf{53.8$\pm$0.9}     \\
\rowcolor{var1light}             & RMDS      & 59.2$\pm$2.1  & 50.0$\pm$1.7     & 49.1$\pm$0.5     & 79.9$\pm$12.1 & 60.9$\pm$1.1           & 50.5$\pm$0.4     \\
\rowcolor{var1light}               & KNN       & 79.4$\pm$1.2  & 53.4$\pm$2.4     & 48.5$\pm$0.8     & 65.4$\pm$8.3  & \textbf{73.3$\pm$0.9}           & 54.0$\pm$0.5     \\
\rowcolor{var1light}               & VIM       & 69.9$\pm$1.4  & 51.2$\pm$1.7     & 46.7$\pm$0.8     & 97.8$\pm$2.4  & 71.1$\pm$1.2           & 54.0$\pm$0.9     \\
\rowcolor{var1light}               & SHE       & 61.5$\pm$0.8  & 50.8$\pm$1.9     & 50.1$\pm$0.1     & 93.9$\pm$5.7  & 56.7$\pm$0.7           & 49.6$\pm$0.2     \\
\rowcolor{var1light}               & KLM       & 56.8$\pm$1.2  & 52.7$\pm$1.8     & 51.7$\pm$0.2     & 79.5$\pm$5.8  & 46.1$\pm$2.0           & 49.3$\pm$0.6     \\
\rowcolor{var1light}               & OpenMax   & 52.1$\pm$1.3  & 47.4$\pm$1.5     & 46.3$\pm$1.3     & 62.7$\pm$10.3 & 66.9$\pm$0.9           & 52.6$\pm$0.5     \\
\rowcolor{var1dark}               & MSP       & 51.2$\pm$1.2  & 47.1$\pm$1.8     & 46.3$\pm$1.3     & 13.4$\pm$7.7  & 66.0$\pm$0.6           & 52.4$\pm$0.4     \\
\rowcolor{var1dark}      MLP          & MLS       & 51.3$\pm$1.4  & 46.8$\pm$1.7     & 46.2$\pm$1.3     & 14.2$\pm$7.6  & 65.9$\pm$0.8           & 52.3$\pm$0.4     \\
\rowcolor{var1dark}                & TempScale & 51.2$\pm$1.2  & 46.8$\pm$1.8     & 46.3$\pm$1.3     & 13.3$\pm$7.7  & 66.0$\pm$0.8           & 52.4$\pm$0.3     \\
\rowcolor{var1dark}                & ODIN      & 51.3$\pm$1.2  & 46.8$\pm$1.8     & 46.3$\pm$1.3     & 13.5$\pm$7.8  & 65.9$\pm$0.8           & 52.5$\pm$0.3     \\
\rowcolor{var1dark}                & EBO       & 51.4$\pm$1.4  & 46.8$\pm$1.7     & 46.2$\pm$1.3     & 14.4$\pm$7.7  & 65.9$\pm$0.8           & 52.3$\pm$0.4     \\
\rowcolor{var1dark}                & GRAM      & 46.7$\pm$1.4  & 46.6$\pm$1.5     & 48.4$\pm$0.6     & 16.7$\pm$11.7 & 53.6$\pm$4.1           & 49.9$\pm$0.1     \\
\rowcolor{var1dark}                & GradNorm  & 50.2$\pm$1.3  & 46.8$\pm$1.8     & 46.4$\pm$1.3     & 12.7$\pm$7.9  & 65.8$\pm$0.8           & 52.5$\pm$0.4     \\
\rowcolor{var1dark}                & ReAct     & 52.5$\pm$1.2  & 46.7$\pm$1.5     & 46.6$\pm$1.4     & 74.3$\pm$19.6 & 66.1$\pm$0.9           & 52.5$\pm$0.4     \\
\rowcolor{var1dark}                & DICE      & 50.6$\pm$1.3  & 46.9$\pm$1.8     & 46.6$\pm$1.3     & 14.7$\pm$8.4  & 65.8$\pm$0.9           & 52.7$\pm$0.8     \\
\rowcolor{var1dark}                & ASH       & 50.6$\pm$1.3  & 46.8$\pm$1.5     & 46.6$\pm$1.7     & 14.0$\pm$7.3  & 65.7$\pm$0.9           & 52.1$\pm$0.4     \\
\midrule
\rowcolor{var2light} \multirow{17}{*}{ResNet}         & MDS       & \textbf{96.9$\pm$0.3}  & \textbf{58.4$\pm$0.6}     & 51.6$\pm$0.7    & \textbf{99.7$\pm$0.1}  & 74.0$\pm$2.4           & 55.4$\pm$0.3     \\
\rowcolor{var2light}               & RMDS      & 45.8$\pm$2.9  & 50.1$\pm$2.5     & 49.9$\pm$0.3     & 79.5$\pm$13.1 & 62.6$\pm$1.4           & 50.8$\pm$0.4     \\
\rowcolor{var2light}               & KNN       & 91.2$\pm$0.9  & 59.2$\pm$1.6     & 50.1$\pm$0.1     & 93.0$\pm$1.9  & 57.7$\pm$2.3           & 54.3$\pm$0.5     \\
\rowcolor{var2light}               & VIM       & 93.5$\pm$1.0  & 56.9$\pm$1.5     & 48.2$\pm$1.1     & 99.5$\pm$0.2  & \textbf{75.0$\pm$1.8}           & \textbf{56.2$\pm$0.4}     \\
\rowcolor{var2light}               & SHE       & 86.7$\pm$1.6  & 50.5$\pm$1.3     & \textbf{51.7$\pm$0.8}     & 99.7$\pm$0.1  & 73.0$\pm$1.6           & 52.8$\pm$0.6     \\
\rowcolor{var2light}               & KLM       & 55.5$\pm$2.6  & 47.5$\pm$1.5     & 51.5$\pm$0.4     & 78.1$\pm$6.7  & 55.2$\pm$1.4           & 47.7$\pm$0.2     \\
\rowcolor{var2light}               & OpenMax   & 64.8$\pm$5.4  & 50.7$\pm$1.3     & 47.0$\pm$1.1     & 65.2$\pm$24.9 & 67.2$\pm$4.1           & 54.3$\pm$0.4     \\
\rowcolor{var2dark}               & MSP       & 63.6$\pm$5.0  & 51.3$\pm$2.4     & 47.1$\pm$0.8     & 19.5$\pm$10.3 & 65.0$\pm$4.3           & 53.9$\pm$0.7     \\
\rowcolor{var2dark}       ResNet        & MLS       & 64.4$\pm$6.0  & 51.9$\pm$2.2     & 46.8$\pm$1.4     & 38.0$\pm$26.5 & 66.5$\pm$4.6           & 54.1$\pm$0.3     \\
\rowcolor{var2dark}               & TempScale & 63.6$\pm$5.0  & 51.3$\pm$2.4     & 47.1$\pm$0.8     & 19.5$\pm$10.3 & 65.0$\pm$4.3           & 53.9$\pm$0.7     \\
\rowcolor{var2dark}               & ODIN      & 63.7$\pm$5.0  & 51.3$\pm$2.4     & 47.1$\pm$0.8     & 19.6$\pm$10.4 & 65.0$\pm$4.4           & 53.9$\pm$0.7     \\
\rowcolor{var2dark}               & EBO       & 64.4$\pm$6.1  & 52.0$\pm$2.2     & 46.8$\pm$1.4     & 38.5$\pm$27.0 & 66.5$\pm$4.6           & 54.2$\pm$0.3     \\
\rowcolor{var2dark}               & GRAM      & 39.3$\pm$2.6  & 48.6$\pm$1.2     & 49.0$\pm$0.9     & 11.3$\pm$5.0  & 44.2$\pm$1.7           & 49.6$\pm$0.1     \\
 \rowcolor{var2dark}              & GradNorm  & 48.1$\pm$5.6  & 49.8$\pm$1.6     & 47.5$\pm$1.1     & 7.5$\pm$4.8   & 48.2$\pm$6.9           & 52.7$\pm$0.6     \\
\rowcolor{var2dark}               & ReAct     & 71.8$\pm$3.9  & 52.7$\pm$1.8     & 47.4$\pm$1.5     & 81.4$\pm$9.1  & 73.9$\pm$2.6           & 55.3$\pm$0.4     \\
\rowcolor{var2dark}               & DICE      & 52.9$\pm$7.9  & 50.5$\pm$1.2     & 47.3$\pm$0.6     & 17.0$\pm$7.6  & 57.7$\pm$7.4           & 55.5$\pm$0.5     \\
 \rowcolor{var2dark}              & ASH       & 68.1$\pm$5.6  & 51.5$\pm$2.4     & 47.2$\pm$0.8     & 40.5$\pm$27.9 & 63.9$\pm$4.3           & 53.9$\pm$0.5     \\
\midrule
\rowcolor{var3light} \multirow{17}{*}{FT-T} & MDS       & 84.1$\pm$4.2  & \textbf{58.5$\pm$2.2}     & 50.8$\pm$1.1     & \textbf{94.2$\pm$0.7}  & 77.8$\pm$2.0          & 50.1$\pm$0.2     \\
\rowcolor{var3light}               & RMDS      & 57.3$\pm$0.9  & 51.6$\pm$1.5     & 48.3$\pm$0.7     & 75.8$\pm$13.5 & 60.6$\pm$4.4           & 50.9$\pm$1.5     \\
\rowcolor{var3light}               & KNN       & 85.5$\pm$3.7  & 55.8$\pm$1.9     & 49.6$\pm$0.2     & 92.0$\pm$0.6  & \textbf{78.6$\pm$3.2}           & 51.7$\pm$1.5     \\
\rowcolor{var3light}               & VIM       & \textbf{85.5$\pm$2.8}  & 57.3$\pm$2.3     & 48.8$\pm$0.1     & 96.8$\pm$0.7  & 79.4$\pm$1.7           & \textbf{53.8$\pm$0.3}     \\
\rowcolor{var3light}               & SHE       & 62.9$\pm$1.3  & 50.5$\pm$1.7     & 50.4$\pm$0.7     & 85.2$\pm$9.1  & 67.5$\pm$0.5           & 47.1$\pm$0.5     \\
\rowcolor{var3light}               & KLM       & 52.0$\pm$2.1  & 51.6$\pm$2.1     & \textbf{51.0$\pm$0.7}     & 74.8$\pm$6.0  & 49.3$\pm$2.6           & 47.3$\pm$1.1     \\
\rowcolor{var3light}               & OpenMax   & 74.3$\pm$2.4  & 48.7$\pm$0.8     & 48.1$\pm$0.5     & 66.5$\pm$8.3  & 77.7$\pm$3.5           & 53.5$\pm$0.2     \\
\rowcolor{var3dark}               & MSP       & 74.3$\pm$2.5  & 48.4$\pm$1.0     & 48.1$\pm$0.5     & 66.1$\pm$7.6  & 77.6$\pm$3.7           & 53.4$\pm$0.2     \\
\rowcolor{var3dark}    FT-T           & MLS       & 74.1$\pm$2.5  & 49.1$\pm$1.1     & 48.0$\pm$0.5     & 65.2$\pm$6.4  & 77.8$\pm$3.2           & 53.3$\pm$0.2     \\
\rowcolor{var3dark}               & TempScale & 74.3$\pm$2.5  & 48.4$\pm$1.0     & 48.1$\pm$0.5     & 66.1$\pm$7.6  & 77.6$\pm$3.7           & 53.4$\pm$0.2     \\
\rowcolor{var3dark}               & ODIN      & 74.4$\pm$2.4  & 48.4$\pm$1.0     & 48.1$\pm$0.5     & 66.3$\pm$7.5  & 77.6$\pm$3.7           & 53.4$\pm$0.2     \\
\rowcolor{var3dark}               & EBO       & 74.1$\pm$2.4  & 49.1$\pm$1.1     & 48.0$\pm$0.5     & 64.4$\pm$5.9  & 77.8$\pm$3.1           & 53.3$\pm$0.3     \\
\rowcolor{var3dark}               & GRAM      & 73.5$\pm$2.3  & 50.8$\pm$2.7     & 48.6$\pm$0.6     & 70.0$\pm$6.8  & 75.2$\pm$5.4           & 51.7$\pm$0.4     \\
\rowcolor{var3dark}               & GradNorm  & 69.5$\pm$5.9  & 50.4$\pm$3.4     & 48.7$\pm$0.8     & 66.4$\pm$6.8  & 62.5$\pm$8.9           & 53.1$\pm$0.5     \\
 \rowcolor{var3dark}              & ReAct     & 73.9$\pm$2.6  & 48.1$\pm$1.1     & 48.2$\pm$0.7     & 64.0$\pm$4.3  & 77.8$\pm$2.6           & 53.4$\pm$0.3     \\
\rowcolor{var3dark}               & DICE      & 74.3$\pm$2.2  & 49.9$\pm$2.8     & 48.1$\pm$0.6     & 77.7$\pm$7.1  & 77.3$\pm$4.1           & 53.2$\pm$0.6     \\
 \rowcolor{var3dark}              & ASH       & 73.0$\pm$3.2  & 52.6$\pm$2.7     & 48.6$\pm$1.3     & 58.0$\pm$8.5  & 75.4$\pm$3.4           & 52.5$\pm$1.1   \\
               \bottomrule
\end{tabular}
\caption{
Comparing OOD detection methods in the \textit{far-OOD} and \textit{near-OOD} settings using eICU and MIMIC-IV datasets, with performance evaluated through AUROC.
The name in the top row shows the dataset considered as ID in the far-OOD experiment and divided for the near-OOD test.
FCU and Eth stand for first care unit and ethnicity as the variables utilized for constructing the ID/OOD sets. 
Results are averaged over 5 runs.
For each model, the best value in each column is shown in bold, and distance-based post-hoc approaches are distinguished with a lighter color.
}
\label{tab:near_far_ood}
\end{table*}

% \usepackage{multirow}
\begin{table*}[]
\footnotesize
%\scriptsize
\centering
\renewcommand{\arraystretch}{0.85}
\setlength{\tabcolsep}{10pt}
\begin{tabular}{c|c|ccc|ccc}
\toprule
\multirow{2}{*}{Model} & \multirow{2}{*}{Method} & \multicolumn{3}{c|}{eICU}                    & \multicolumn{3}{c}{MIMIC-IV}                 \\
                       &                         & $\mathcal{F}$=10         & $\mathcal{F}$=100        & $\mathcal{F}$=1000        & $\mathcal{F}$=10         & $\mathcal{F}$=100         & $\mathcal{F}$=1000        \\
\midrule
%\multirow{6}{*}{-} 
\rowcolor{var4} & AE                      & 80.5$\pm$1.3 & 88.4$\pm$0.9 & 90.0$\pm$0.7  & 76.4$\pm$1.6 & 83.9$\pm$2.1  & 86.6$\pm$2.1  \\
\rowcolor{var4}                       & VAE                     & 80.0$\pm$1.3 & 88.3$\pm$0.9 & 89.9$\pm$0.7  & 76.4$\pm$1.6 & 83.8$\pm$2.1  & 86.6$\pm$2.1  \\
\rowcolor{var4}  & HI-VAE                    & 50.0$\pm$0.1 & 50.0$\pm$0.1 & 50.1$\pm$0.2  & 50.59$\pm$0.8 & 51.3$\pm$1.4  & 52.0$\pm$2.0  \\
\rowcolor{var4} Density & Flow                    & 70.2$\pm$2.5 & 82.1$\pm$2.5 & 87.7$\pm$1.4  & 53.8$\pm$1.8 & 65.0$\pm$3.0  & 75.7$\pm$2.5  \\
\rowcolor{var4} Based & PPCA                    & 80.7$\pm$1.3 & 88.3$\pm$0.9 & 89.7$\pm$0.8  & 76.9$\pm$1.5 & 84.0$\pm$2.1  & 86.6$\pm$2.0  \\
\rowcolor{var4}                       & LOF                     & \textbf{84.4$\pm$1.3} & \textbf{89.4$\pm$0.8} & \textbf{90.5$\pm$0.7}  & \textbf{78.4$\pm$1.5} & \textbf{84.7$\pm$2.1}  & \textbf{86.9$\pm$1.9}  \\
\rowcolor{var4}                       & DUE                     & 63.9$\pm$1.6 & 80.5$\pm$1.3 & 88.7$\pm$0.8  & 60.3$\pm$2.2 & 76.0$\pm$1.9  & 83.0$\pm$2.0  \\
\midrule
\rowcolor{var1light} \multirow{17}{*}{MLP}                    & MDS                     & \textbf{68.5$\pm$2.4} & \textbf{82.8$\pm$2.0} & 89.2$\pm$2.0  & \textbf{69.3$\pm$2.4} & \textbf{80.8$\pm$1.6}  & \textbf{84.7$\pm$1.8}  \\
\rowcolor{var1light}                       & RMDS                    & 60.8$\pm$1.0 & 75.2$\pm$2.0 & 85.8$\pm$2.0  & 52.7$\pm$4.1 & 64.5$\pm$6.9  & 76.4$\pm$2.8  \\
\rowcolor{var1light}                       & KNN                     & 60.3$\pm$1.5 & 68.0$\pm$2.0 & 70.9$\pm$2.3  & 59.1$\pm$4.9 & 67.2$\pm$8.9  & 70.7$\pm$10.0 \\
\rowcolor{var1light}                       & VIM                     & 48.5$\pm$2.0 & 47.2$\pm$3.1 & 46.0$\pm$4.9  & 56.8$\pm$7.9 & 63.5$\pm$13.7 & 66.6$\pm$15.1 \\
\rowcolor{var1light}                       & SHE                     & 62.5$\pm$2.8 & 77.8$\pm$1.6 & \textbf{89.8$\pm$0.1}  & 62.3$\pm$3.0 & 77.8$\pm$2.2  & 83.6$\pm$1.5  \\
\rowcolor{var1light}                       & KLM                     & 60.2$\pm$1.4 & 69.6$\pm$1.5 & 78.1$\pm$1.4  & 55.4$\pm$2.0 & 66.9$\pm$1.6  & 74.0$\pm$1.5  \\
\rowcolor{var1light}                       & OpenMax                 & 52.6$\pm$2.0 & 65.2$\pm$2.5 & 77.4$\pm$2.5  & 48.4$\pm$5.3 & 54.0$\pm$5.1  & 69.4$\pm$5.6  \\
\rowcolor{var1dark}                       & MSP                     & 40.5$\pm$2.2 & 27.1$\pm$3.0 & 14.2$\pm$2.4  & 45.7$\pm$4.8 & 31.7$\pm$5.5  & 21.8$\pm$1.3  \\
\rowcolor{var1dark} MLP                       & MLS                     & 40.4$\pm$2.4 & 27.0$\pm$3.4 & 13.8$\pm$2.8  & 45.6$\pm$4.1 & 32.1$\pm$4.0  & 24.9$\pm$3.7  \\
\rowcolor{var1dark}                        & TempScale               & 40.4$\pm$2.2 & 27.0$\pm$3.1 & 14.0$\pm$2.4  & 45.7$\pm$4.8 & 31.6$\pm$5.5  & 21.7$\pm$1.3  \\
\rowcolor{var1dark}                        & ODIN                    & 40.5$\pm$2.2 & 27.0$\pm$3.1 & 14.0$\pm$2.4  & 45.7$\pm$4.8 & 31.7$\pm$5.5  & 21.8$\pm$1.3  \\
\rowcolor{var1dark}                        & EBO                     & 40.4$\pm$2.4 & 27.0$\pm$3.4 & 13.8$\pm$2.8  & 45.5$\pm$3.9 & 32.0$\pm$4.0  & 24.9$\pm$3.8  \\
\rowcolor{var1dark}                       & GRAM                    & 38.7$\pm$2.0 & 25.2$\pm$2.6 & 12.7$\pm$2.1  & 47.0$\pm$0.7 & 32.8$\pm$3.0  & 20.4$\pm$2.2  \\
\rowcolor{var1dark}                        & GradNorm                & 40.2$\pm$2.1 & 26.4$\pm$2.9 & 13.2$\pm$2.4  & 42.6$\pm$2.4 & 26.1$\pm$1.1  & 18.8$\pm$1.4  \\
\rowcolor{var1dark}                        & ReAct                   & 45.0$\pm$1.6 & 37.3$\pm$2.7 & 31.3$\pm$2.5  & 49.0$\pm$5.7 & 46.6$\pm$12.4 & 46.7$\pm$14.5 \\
\rowcolor{var1dark}                        & DICE                    & 40.9$\pm$2.7 & 28.9$\pm$3.5 & 16.7$\pm$2.9  & 44.8$\pm$3.6 & 31.4$\pm$2.7  & 22.0$\pm$1.1  \\
\rowcolor{var1dark}                        & ASH                     & 40.4$\pm$2.4 & 27.3$\pm$3.1 & 14.3$\pm$2.6  & 45.3$\pm$4.1 & 32.1$\pm$4.3  & 25.1$\pm$3.7  \\
\midrule
\rowcolor{var2light} \multirow{17}{*}{ResNet}                 & MDS                     & \textbf{74.4$\pm$2.0} & \textbf{88.9$\pm$1.7} & 91.3$\pm$1.4  & \textbf{72.4$\pm$1.1} & \textbf{81.8$\pm$0.8}  & \textbf{85.4$\pm$1.0}  \\
\rowcolor{var2light}                       & RMDS                    & 51.7$\pm$0.6 & 59.8$\pm$2.5 & 78.7$\pm$2.4  & 46.2$\pm$1.5 & 57.6$\pm$2.7  & 73.9$\pm$1.2  \\
\rowcolor{var2light}                       & KNN                     & 69.5$\pm$2.1 & 84.8$\pm$2.2 & 87.9$\pm$2.2  & 67.5$\pm$1.5 & 79.1$\pm$1.2  & 83.7$\pm$0.8  \\
\rowcolor{var2light}                       & VIM                     & 72.2$\pm$2.5 & 88.7$\pm$1.8 & 91.5$\pm$1.4  & 68.8$\pm$1.3 & 80.2$\pm$1.2  & 84.5$\pm$1.1  \\
\rowcolor{var2light}                       & SHE                     & 68.2$\pm$0.7 & 86.6$\pm$0.5 & \textbf{91.5$\pm$0.2}  & 67.2$\pm$1.4 & 80.0$\pm$0.9  & 84.5$\pm$0.7  \\
\rowcolor{var2light}                       & KLM                     & 56.6$\pm$0.8 & 69.4$\pm$1.7 & 80.5$\pm$1.9  & 53.2$\pm$0.9 & 65.4$\pm$0.5  & 75.3$\pm$0.8  \\
\rowcolor{var2light}                       & OpenMax                 & 57.5$\pm$0.7 & 66.4$\pm$4.3 & 77.4$\pm$2.9  & 55.2$\pm$1.0 & 61.6$\pm$1.4  & 61.7$\pm$11.4 \\
 \rowcolor{var2dark}                       & MSP                     & 49.4$\pm$2.2 & 35.1$\pm$2.1 & 16.8$\pm$1.0  & 52.2$\pm$1.1 & 38.0$\pm$0.7  & 23.1$\pm$0.9  \\
 \rowcolor{var2dark} ResNet                       & MLS                     & 49.0$\pm$1.7 & 34.7$\pm$2.0 & 18.5$\pm$2.8  & 52.5$\pm$1.9 & 40.5$\pm$7.0  & 31.0$\pm$8.8  \\
 \rowcolor{var2dark}                       & TempScale               & 49.4$\pm$2.2 & 35.1$\pm$2.1 & 16.8$\pm$1.0  & 52.2$\pm$1.1 & 38.0$\pm$0.7  & 23.1$\pm$0.9  \\
 \rowcolor{var2dark}                       & ODIN                    & 49.4$\pm$2.2 & 35.2$\pm$2.1 & 16.8$\pm$1.0  & 52.3$\pm$1.1 & 38.1$\pm$0.7  & 23.1$\pm$0.9  \\
 \rowcolor{var2dark}                       & EBO                     & 49.0$\pm$1.6 & 34.5$\pm$2.1 & 18.5$\pm$2.8  & 52.4$\pm$2.1 & 40.6$\pm$7.4  & 31.0$\pm$9.0  \\
 \rowcolor{var2dark}                       & GRAM                    & 37.0$\pm$3.1 & 17.7$\pm$3.0 & 9.1$\pm$1.6   & 46.6$\pm$1.2 & 29.0$\pm$0.7  & 18.9$\pm$0.9  \\
 \rowcolor{var2dark}                       & GradNorm                & 40.1$\pm$2.4 & 20.4$\pm$1.0 & 9.7$\pm$1.6   & 44.9$\pm$1.0 & 25.7$\pm$0.9  & 17.2$\pm$0.3  \\
  \rowcolor{var2dark}                      & ReAct                   & 59.4$\pm$1.5 & 70.4$\pm$3.5 & 74.8$\pm$3.5  & 58.7$\pm$2.0 & 67.2$\pm$4.1  & 70.2$\pm$4.4  \\
  \rowcolor{var2dark}                      & DICE                    & 42.6$\pm$2.0 & 24.6$\pm$1.2 & 10.8$\pm$0.9  & 50.6$\pm$1.6 & 36.8$\pm$1.6  & 23.1$\pm$1.5  \\
   \rowcolor{var2dark}                     & ASH                     & 49.8$\pm$0.3 & 36.7$\pm$2.4 & 20.1$\pm$2.9  & 52.3$\pm$1.7 & 40.9$\pm$5.2  & 31.1$\pm$8.5  \\
\midrule
 \rowcolor{var3light} \multirow{17}{*}{FT-T}         & MDS                     & \textbf{59.9$\pm$1.4} & \textbf{79.5$\pm$1.4} & 87.5$\pm$0.9  & \textbf{59.7$\pm$1.1} & \textbf{76.0$\pm$1.2}  & \textbf{81.8$\pm$1.4}  \\
 \rowcolor{var3light}                      & RMDS                    & 51.5$\pm$1.3 & 57.8$\pm$7.4 & 64.0$\pm$13.0 & 49.0$\pm$0.9 & 55.0$\pm$2.0  & 67.0$\pm$5.1  \\
  \rowcolor{var3light}                     & KNN                     & 57.3$\pm$1.4 & 75.4$\pm$2.2 & 86.5$\pm$1.3  & 57.6$\pm$1.2 & 74.6$\pm$1.0  & 81.4$\pm$1.3  \\
  \rowcolor{var3light}                     & VIM                     & 57.9$\pm$1.6 & 77.6$\pm$1.3 & \textbf{88.3$\pm$0.7}  & 57.4$\pm$2.5 & 74.3$\pm$1.2  & 80.9$\pm$1.2  \\
  \rowcolor{var3light}                     & SHE                     & 55.7$\pm$1.3 & 71.2$\pm$2.9 & 80.4$\pm$1.6  & 53.2$\pm$0.7 & 67.6$\pm$1.6  & 80.8$\pm$2.4  \\
  \rowcolor{var3light}                     & KLM                     & 54.1$\pm$0.8 & 63.1$\pm$1.1 & 72.1$\pm$4.2  & 52.8$\pm$0.4 & 60.8$\pm$1.1  & 61.8$\pm$6.9  \\
   \rowcolor{var3light}                    & OpenMax                 & 51.0$\pm$0.7 & 56.1$\pm$2.7 & 71.4$\pm$3.2  & 48.8$\pm$0.9 & 54.1$\pm$2.6  & 66.4$\pm$3.8  \\
  \rowcolor{var3dark}                     & MSP                     & 50.9$\pm$0.6 & 55.8$\pm$2.7 & 71.3$\pm$3.1  & 48.4$\pm$0.8 & 49.6$\pm$1.5  & 62.9$\pm$2.9  \\
\rowcolor{var3dark}  FT-T                      & MLS                     & 50.9$\pm$0.7 & 55.8$\pm$3.1 & 70.8$\pm$3.1  & 48.4$\pm$0.9 & 49.5$\pm$2.7  & 60.5$\pm$5.6  \\
 \rowcolor{var3dark}                       & TempScale               & 50.9$\pm$0.6 & 55.8$\pm$2.7 & 71.3$\pm$3.1  & 48.4$\pm$0.8 & 49.6$\pm$1.5  & 62.9$\pm$2.9  \\
\rowcolor{var3dark}                        & ODIN                    & 50.9$\pm$0.6 & 55.9$\pm$2.7 & 71.4$\pm$3.1  & 48.4$\pm$0.8 & 49.7$\pm$1.5  & 63.1$\pm$2.9  \\
 \rowcolor{var3dark}                       & EBO                     & 50.9$\pm$0.7 & 55.5$\pm$3.1 & 70.2$\pm$3.2  & 48.5$\pm$1.0 & 49.0$\pm$3.4  & 57.3$\pm$7.4  \\
 \rowcolor{var3dark}                       & GRAM                    & 48.9$\pm$0.1 & 50.1$\pm$0.8 & 62.5$\pm$3.0  & 47.8$\pm$1.3 & 44.0$\pm$2.7  & 49.6$\pm$7.6  \\
 \rowcolor{var3dark}                       & GradNorm                & 52.0$\pm$0.8 & 59.5$\pm$2.9 & 73.6$\pm$2.9  & 48.8$\pm$0.9 & 51.4$\pm$1.7  & 64.2$\pm$2.7  \\
 \rowcolor{var3dark}                       & ReAct                   & 50.9$\pm$0.8 & 55.7$\pm$3.4 & 69.9$\pm$3.2  & 48.6$\pm$1.1 & 49.7$\pm$3.4  & 57.4$\pm$7.1  \\
  \rowcolor{var3dark}                      & DICE                    & 52.2$\pm$1.5 & 62.8$\pm$2.5 & 75.6$\pm$1.4  & 50.2$\pm$0.9 & 64.5$\pm$1.4  & 76.5$\pm$1.6  \\
 \rowcolor{var3dark}                       & ASH                     & 50.1$\pm$1.4 & 54.6$\pm$2.1 & 69.7$\pm$1.3  & 48.7$\pm$0.9 & 49.9$\pm$2.8  & 59.2$\pm$5.6 \\
                       \bottomrule
\end{tabular}
\caption{
Comparing OOD detection methods in the \textit{synthesized-OOD} setting using eICU and MIMIC-IV datasets, with performance evaluated through AUROC.
In this setting, one feature is multiplied by a specific factor $\mathcal{F}$ to construct an OOD sample. For each model and factor we display the average performance, calculated over 100 resampling of the feature and 5 different initializations of the model. 
%\uparrow$ means higher is better and $\downarrow$ means lower is better. 
For each model, the best value in each column is shown in bold, and distance-based post-hoc approaches are distinguished with a lighter color.
}
\label{tab:synthesized_ood}
\end{table*}

\begin{figure*}[t]
    \centering
    % \vskip -10pt
    \includegraphics[scale=0.4]{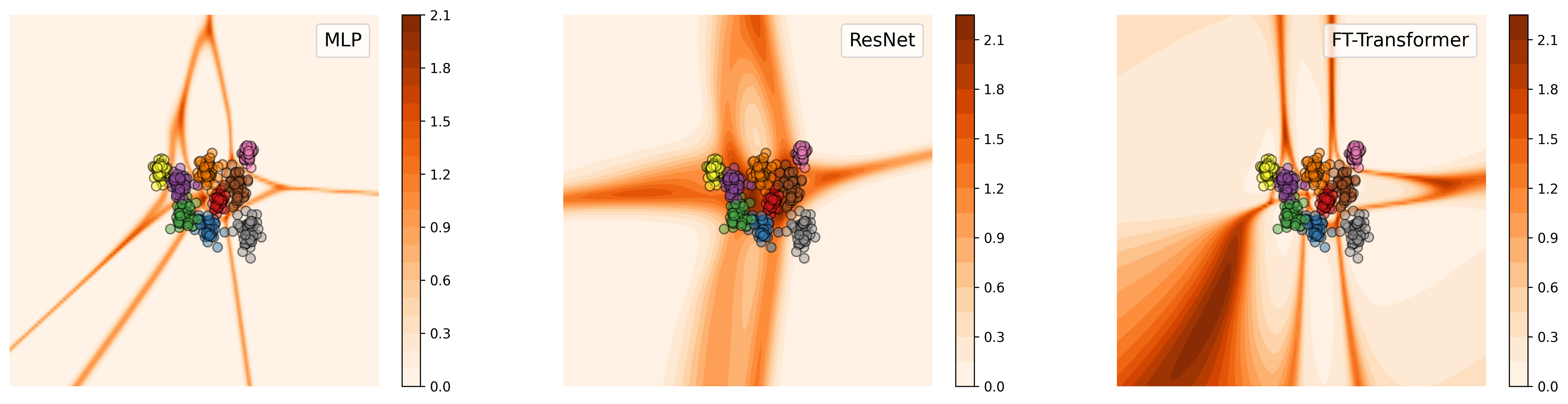}
    \caption{Depiction of confidence scores for different architectures. High confidence (low entropy) is represented in light orange. MLP and ResNet exhibit regions of confidence extending far away from the ID data. FT-Transformer mitigates this over-confidence but does not solve it.
    % Comparison of over-confidence problem in different architectures using the entropy of the softmax output in a multi-class classification toy example. Higher entropy is equivalent to lower confidence. FT-Transformer mitigates the over-confidence issue of MLP and ResNet but does not solve it.
    }
    %\vskip -5pt
    \label{fig:over_confidence}
\end{figure*}

\subsection{Synthesized-OOD}

Results for the synthesized-OOD setting are displayed in Table \ref{tab:synthesized_ood}. 
It is expected that increasing the scaling factor ($\mathcal{F}$) in this setting facilitates the detection of OODs. However, over-confidence hampers OOD detection, causing certain methods to exhibit a totally opposite behavior with MLP and ResNet architectures, on both datasets. In this case, even the diversity in eICU does not prevent over-confidence. Similar to the far-OOD scenario, FT-Transformer seems to solve this issue, as increasing $\mathcal{F}$ results in an improved performance for this architecture.

To compare different approaches, density-based methods like AE, VAE, PPCA, and LOF demonstrate better performance than post-hoc ones when $\mathcal{F}=10$. However, this gap in performance is reduced with increasing $\mathcal{F}$ to the extent that methods like MDS, VIM, and SHE applied on ResNet are outperforming density-based ones on the eICU dataset with $\mathcal{F}=1000$.

\subsection{Over-confidence}

The results above showed over-confidence for the MLP and ResNet architectures, whereas the FT-Transformer appeared as a solution to this problem. 
For a more visual exploration of this issue, we employ a classification toy example. In this example, each of the predictive architectures is trained on a multi-class classification task with 2D samples. Next, the entropy of the model's softmax output is plotted for a wide array of inputs. Plots are shown in Fig. \ref{fig:over_confidence}, with a lighter color indicating more confidence (i.e. low entropy in the softmax output). As depicted in this figure, MLP and ResNet confidence is increased by getting further away from the ID data. This observation validates the presence of over-confidence in both the MLP and ResNet models. Conversely, in the FT-Transformer, confidence increases along certain directions and decreases in others. This suggests that the transformer can mitigate the over-confidence in some directions, but does not solve it entirely.

\section{Discussion}
\label{sec:Discussion}

%Throughout this section, we collectively examine the findings. 
According to our results, when OODs are far from ID data, there exist methods that can effectively perform OOD detection. However, OODs near to ID data such as multiplication by $\mathcal{F}=10$ or in a near-OOD setting are still challenging to spot. While the poor results in the near-OOD scenarios might be due to a large overlap with the ID data, when multiplying features with a scaling factor one can ensure that there is a clear difference between such OOD and ID data. 
This points to the fact that there is still ample room for improvement in detecting near OODs.

Delving into OOD detectors, density-based methods, particularly AE, VAE, PPCA, and Flow, exhibit consistent good performance across different settings and datasets. 
Within the post-hoc methods, they perform poorly in some cases, but they improve substantially and become competitive with the density-based ones when used in conjunction with the distance-based mechanisms. For example, MDS applied on ResNet can even marginally outperform density-based methods in some cases, such as the experiment with a scaling factor of $\mathcal{F}=1000$ on the eICU dataset. This nuances  previous claims that post-hoc methods generally exhibit poor performance \citep{Ulmer2020Trust, zadorozhny2022out}.  

%MDS is better than others and it is competitive with the density-based ones when used in conjunction with ResNet. Furthermore, it can also marginally outperform them in some cases,  such as far-OOD scenario with multiplication by $\mathcal{F}=1000$ on the eICU dataset. After MDS, distance-based approaches like KNN, VIM, and SHE demonstrate good performance in general. 

%It's important to highlight that there might be an inclination to label ID samples that the predictive model is uncertain about as OOD to avoid wrong predictions in the ID set. This adjustment would cause an improvement in the performance of OOD detectors that rely on analogous novelty score algorithms as the primary task the model is trained on. This has been shown in the Appendix \ref{apd:missclass}. Consequently, opting for such detectors in practical applications would be beneficial.

To compare different prediction models, ResNet combined with distance-based detectors demonstrates better results than MLP and FT-Transformer.
On the other hand, MLP and ResNet suffer from over-confidence, causing certain detectors to perform even worse than a random classifier. 
This aligns with what was highlighted in prior studies  \citep{hein2019relu, ulmer2021know}. Our numerical results suggest that FT-Transformer could be a solution to this problem, however, our simple example with toy data showcases that transformers do not completely eliminate over-confidence.

This benchmark is built on two intensive care datasets, which are highly granular. Hence, caution should be exercised when transporting these findings to alternative healthcare tabular datasets with different characteristics. To facilitate the extension of this benchmark, we provided a modular implementation allowing for the addition of new datasets and methods to the experiments.

% the OOD detection methods are implemented modular and can be simply adapted within any other tabular dataset. 

%These points should be taken into account when building OOD detection models for healthcare applications on tabular data.

%Accordingly, there is a trade-off between over-confidence and OOD detection performance in ResNet and FT-Transformer, which should be taken into account when selecting these architectures for the primary task. 

%\acks{Acknowledgements go here.}

\bibliography{jmlr-sample}

\newpage
\appendix
\section{Clinical Variables}\label{apd:variables}
Clinical variables selected for each of the eICU and MIMIC-IV datasets are provided in Table \ref{tab:variables}. These variables are mostly selected as suggested in prior works \citep{Ulmer2020Trust, spathis2022looking, DKL_Miguel2021}. Additionally, we have excluded some time-dependent variables in MIMIC-IV since they are not recorded for most patients. Adding these variables would result in a lot of NaN values. Note that when a dataset is considered as an OOD set for the other one, only available variables for both datasets are included in that experiment.

\begin{table}[h]
\centering
\begin{tabular}{ll}
\toprule
eICU                  & MIMIC-IV                 \\ \midrule
\multicolumn{2}{l}{\textbullet{} \textit{Time-dependent}}             \vspace{3pt}  \\
pH                    & pH                       \\
Temperature (c)        & Temperature              \\
Respiratory Rate      & Respiratory rate         \\
O2 Saturation         & Oxygen saturation        \\
MAP (mmHg)            & Mean blood pressure      \\
Heart Rate            & Heart Rate               \\
glucose               & Glucose                  \\
GCS Total             & -                \\
Motor                 & -                 \\
Eyes                  & -                \\
Verbal                & -                 \\
FiO2                  & -               \\
Invasive BP Diastolic & Diastolic blood pressure \\
Invasive BP Systolic  & Systolic blood pressure  \\ \midrule
\multicolumn{2}{l}{\textbullet{} \textit{Time-independent}}           \vspace{3pt}  \\
gender                & gender                   \\
age                   & age                      \\
ethnicity             & -               \\
admissionheight       & -                \\
admissionweight       & -               \\
-              & admission\_type          \\
-             & first\_careunit     \\ \bottomrule    
\end{tabular}
\caption{Clinical time-dependent and time-independent variables used for each of eICU and MIMIC-IV datasets. Corresponding variables are put in the same row, with a dash indicating that the variable is not included in a dataset.}
\label{tab:variables}
\end{table}

\section{Mortality Prediction}\label{apd:mortality}

Post-hoc OOD detectors are utilized alongside a prediction model. It is essential that this prediction model is trained well on the ID data \citep{vaze2021open}. Otherwise, the detector would not be able to distinguish IDs and OODs. 

Following the guidelines in \cite{gorishniy2021revisiting}, we have trained the prediction models for $10$ epochs with a batch size equal to 64 and AdamW \citep{loshchilov2018decoupled} optimizer with a learning rate equal to $1e\text{-}3$. 

To ensure that our prediction models are trained well, we have measured the AUROC of the mortality prediction task on the test set. Results are presented in Table \ref{tab:mortality}, indicating that models are able to solve the task well. Also, it demonstrates that transformer architecture achieves better performance on this task than MLP and ResNet across both datasets.

Additionally, according to the results in Table \ref{tab:near_far_ood} and \ref{tab:synthesized_ood}, MLP and ResNet in conjunction with distance-based post-hoc detectors can outperform transformers in many cases. This suggests that superior performance as a closed-set classifier on ID data does not necessarily translate to better OOD detection, challenging the claim in \cite{vaze2021open} that better closed-set classifiers lead to improved OOD detection performance.

\begin{table}[h]
\centering
\begin{tabular}{l|cc}
               & eICU         & MIMIC-IV     \\ \midrule
MLP            & 85.7$\pm$1.6 & 75.6$\pm$1.0 \\
ResNet         & 87.9$\pm$0.5 & 77.3$\pm$1.2 \\
FT-Transformer & 89.1$\pm$0.4 & 80.1$\pm$0.3
\end{tabular}
\caption{AUROC of mortality prediction task for MLP, ResNet, and Ft-Transformer architectures on eICU and MIMIC-IV. Results are averaged over 5 runs.}
\label{tab:mortality}
\end{table}

\section{Other Near-OOD Results}\label{apd:near_ood}

Near-OOD results based on age (older than 70 as ID), ethnicity (Caucasian as ID), and first care unit (CVICU as ID) are provided in Table \ref{tab:near_far_ood}. To explore this with other different variables, results for alternate variables are presented in table \ref{tab:other_near}. The variables used in this table are gender (females as ID) and ethnicity (African Americans as ID) for eICU, and gender (females as ID) and admission type (surgical on the same day of admission as ID) for MIMIC-IV. 

In eICU, results for both variables indicate an almost random AUC (50\%) for all the methods. In MIMIC-IV, gender shows a similar behavior, while for admission type, certain approaches like MDS and AE exhibit marginal superiority over the rest. These results align well with the observations from previous variables in the table \ref{tab:near_far_ood}. First of all, the wide-ranging diversity present in the eICU dataset poses challenges in identifying nearly out-of-distribution sets. Additionally, in scenarios where the difficulty level is not excessively high to the point of random performance, methods like AE and MDS emerge as competitive options. Lastly, the issue of detecting near-OOD samples is still challenging.

% \usepackage{multirow}
\begin{table*}[]
\footnotesize
%\scriptsize
\centering
\renewcommand{\arraystretch}{0.85}
\setlength{\tabcolsep}{5.0pt}
\begin{tabular}{c|c|cccc|cccc}
\toprule
\multirow{3}{*}{Model}   & \multirow{3}{*}{Method} & \multicolumn{4}{c|}{eICU}                                                         & \multicolumn{4}{c}{MIMIC-IV}                                                     \\
                         &                         & \multicolumn{2}{c}{Near-OOD(Gender)} & \multicolumn{2}{c|}{Near-OOD(Ethnicity)} & \multicolumn{2}{c}{Near-OOD(Gender)} & \multicolumn{2}{c}{Near-OOD(Admission)} \\
                         &                         & AUROC             & FPR95             & AUROC              & FPR95               & AUROC             & FPR95             & AUROC        & \multicolumn{1}{c}{FPR95} \\ 
                         \midrule
\rowcolor{var4}  \multicolumn{1}{l|}{}     & AE                      & 51.0$\pm$0.1      & 95.0$\pm$0.2      & 47.4$\pm$0.4       & 95.9$\pm$0.7        & 50.2$\pm$0.0      & 94.4$\pm$0.1      & 63.1$\pm$1.3 & 94.0$\pm$0.1              \\
\rowcolor{var4} \multicolumn{1}{l|}{}     & VAE                     & 50.7$\pm$0.2      & 95.0$\pm$0.2      & 47.3$\pm$0.4       & 95.9$\pm$0.7        & 50.2$\pm$0.0      & 94.4$\pm$0.1      & \textbf{63.2$\pm$1.2} & 93.9$\pm$0.1              \\
\rowcolor{var4}  & HI-VAE                    & 42.9$\pm$ 0.4 & 100.0$\pm$0.0 & 46.9$\pm$1.6  & 97.2$\pm$1.5 & 50.2$\pm$1.1 & 94.8$\pm$0.4 & 59.0$\pm$18.0 & \textbf{82.1$\pm$12.8} \\
\rowcolor{var4} Density                  & Flow                    & 48.6$\pm$0.3      & 96.3$\pm$0.6      & 48.3$\pm$0.4       & 95.9$\pm$0.1        & 48.2$\pm$0.4      & 95.1$\pm$0.5      & 44.0$\pm$4.2 & 93.9$\pm$0.2              \\
\rowcolor{var4} Based                    & PPCA                    & 51.2$\pm$0.2      & 95.1$\pm$0.1      & 48.1$\pm$0.4       & 95.9$\pm$0.2        & 50.4$\pm$0.2      & 94.4$\pm$0.0      & 61.4$\pm$0.5 & 92.1$\pm$1.2              \\
\rowcolor{var4} \multicolumn{1}{l|}{}     & LOF                     & \textbf{53.1$\pm$0.3}      & \textbf{94.2$\pm$0.1}      & \textbf{50.2$\pm$0.6}       & \textbf{94.4$\pm$0.1}        & \textbf{50.6$\pm$0.0}      & \textbf{94.1$\pm$0.3}      & 60.5$\pm$0.5 & 92.1$\pm$0.8              \\
\rowcolor{var4} \multicolumn{1}{l|}{}     & DUE                     & 50.9$\pm$0.6      & 94.5$\pm$0.3      & 48.7$\pm$0.3       & 95.0$\pm$0.2        & 49.9$\pm$0.1      & 95.4$\pm$0.5      & 57.1$\pm$1.0 & 93.7$\pm$1.5              \\
\midrule
\rowcolor{var1light} \multirow{17}{*}{MLP}    & MDS                     & 48.8$\pm$0.1      & 94.9$\pm$0.0      & 45.4$\pm$0.2       & 95.4$\pm$0.4        & \textbf{51.5$\pm$0.7}      & 94.4$\pm$0.2      & 60.8$\pm$0.2 & 91.5$\pm$1.0              \\
  \rowcolor{var1light}                       & RMDS                    & 50.9$\pm$0.3      & 94.6$\pm$0.3      & 50.5$\pm$0.8       & 94.8$\pm$0.3        & 48.8$\pm$0.8      & 95.3$\pm$0.6      & 56.1$\pm$1.4 & 91.1$\pm$1.8              \\
 \rowcolor{var1light}                        & KNN                     & 52.1$\pm$0.4      & 94.4$\pm$0.3      & 54.5$\pm$0.6       & 94.5$\pm$0.3        & 48.3$\pm$1.0      & 95.0$\pm$0.1      & \textbf{61.6$\pm$0.6} & 93.5$\pm$1.2              \\
   \rowcolor{var1light}                      & VIM                     & 52.2$\pm$0.2      & 94.5$\pm$0.1      & 54.9$\pm$0.7       & 95.2$\pm$0.4        & 48.0$\pm$0.7      & 94.9$\pm$0.2      & 60.5$\pm$0.4 & 91.3$\pm$1.1              \\
   \rowcolor{var1light}                      & SHE                     & 49.0$\pm$0.1      & 95.9$\pm$0.3      & 49.8$\pm$0.8       & 97.3$\pm$0.1        & 50.8$\pm$0.2      & \textbf{93.5$\pm$0.8}      & 54.0$\pm$1.0 & 92.9$\pm$0.9              \\
    \rowcolor{var1light}                     & KLM                     & 48.8$\pm$0.1      & 94.7$\pm$0.2      & 45.1$\pm$0.6       & 94.1$\pm$0.2        & 51.2$\pm$0.2      & 95.4$\pm$0.1      & 48.5$\pm$0.5 & 91.9$\pm$1.3              \\
      \rowcolor{var1light}                   & OpenMax                 & 52.3$\pm$0.2      & 95.0$\pm$0.1      & 56.5$\pm$0.7       & 94.3$\pm$0.1        & 47.1$\pm$0.7      & 95.5$\pm$0.1      & 57.7$\pm$1.3 & 91.2$\pm$0.6              \\
    \rowcolor{var1dark}                     & MSP                     & 52.2$\pm$0.2      & 94.7$\pm$0.2      & 56.8$\pm$0.7       & 94.0$\pm$0.0        & 47.1$\pm$0.8      & 95.5$\pm$0.3      & 57.2$\pm$0.9 & 92.3$\pm$1.9              \\
\rowcolor{var1dark} MLP                         & MLS                     & 52.3$\pm$0.2      & 94.6$\pm$0.1      & 56.9$\pm$0.7       & 94.0$\pm$0.1        & 47.0$\pm$0.8      & 95.2$\pm$0.1      & 57.0$\pm$1.0 & 91.9$\pm$2.1              \\
 \rowcolor{var1dark}                        & TempScale               & 52.2$\pm$0.2      & 94.7$\pm$0.2      & 56.9$\pm$0.7       & 94.0$\pm$0.0        & 47.1$\pm$0.8      & 95.5$\pm$0.3      & 57.0$\pm$0.9 & 92.3$\pm$1.9              \\
   \rowcolor{var1dark}                      & ODIN                    & 52.2$\pm$0.2      & 94.7$\pm$0.1      & 56.8$\pm$0.7       & 94.0$\pm$0.1        & 47.1$\pm$0.8      & 95.3$\pm$0.2      & 57.0$\pm$1.0 & 92.3$\pm$1.8              \\
   \rowcolor{var1dark}                      & EBO                     & 52.3$\pm$0.3      & 94.9$\pm$0.1      & 56.9$\pm$0.7       & 94.1$\pm$0.1        & 47.0$\pm$0.8      & 95.3$\pm$0.0      & 57.0$\pm$1.0 & 91.8$\pm$2.1              \\
    \rowcolor{var1dark}                     & GRAM                    & 51.6$\pm$0.6      & 100.0$\pm$0.0     & 55.0$\pm$1.8       & 100.0$\pm$0.0       & 50.0$\pm$0.1      & 100.0$\pm$0.0     & 49.5$\pm$0.8 & 100.0$\pm$0.0             \\
    \rowcolor{var1dark}                     & GradNorm                & 52.2$\pm$0.2      & 94.5$\pm$0.2      & 56.8$\pm$0.7       & 93.8$\pm$0.2        & 47.1$\pm$0.7      & 95.4$\pm$0.3      & 57.0$\pm$0.9 & 90.8$\pm$1.1              \\
    \rowcolor{var1dark}                     & ReAct                   & 51.9$\pm$0.3      & 94.9$\pm$0.1      & 56.8$\pm$0.7       & 94.1$\pm$0.1        & 47.2$\pm$0.8      & 95.3$\pm$0.0      & 57.0$\pm$1.2 & 91.8$\pm$2.1              \\
     \rowcolor{var1dark}                    & DICE                    & \textbf{52.3$\pm$0.2 }     & \textbf{94.2$\pm$0.1}      & 56.7$\pm$0.8       & 94.6$\pm$0.5        & 47.1$\pm$0.5      & 95.5$\pm$0.3      & 57.0$\pm$1.0 & \textbf{90.3$\pm$1.0}              \\
      \rowcolor{var1dark}                   & ASH                     & 52.1$\pm$0.3      & 94.8$\pm$0.1      & \textbf{56.9$\pm$0.4}       & \textbf{94.0$\pm$0.2}        & 47.3$\pm$1.0      & 95.2$\pm$0.1      & 56.7$\pm$2.2 & 90.6$\pm$1.0              \\
\midrule
\rowcolor{var2light} \multirow{17}{*}{ResNet} & MDS                     & 50.4$\pm$0.4      & 95.2$\pm$0.2      & 47.8$\pm$0.2       & 96.2$\pm$0.1        & 50.6$\pm$0.1      & 94.2$\pm$0.4      & \textbf{62.2$\pm$0.3} & 92.2$\pm$1.0              \\
 \rowcolor{var2light}                        & RMDS                    & 50.5$\pm$0.5      & 94.8$\pm$0.2      & 50.7$\pm$0.6       & 94.9$\pm$0.3        & 50.2$\pm$0.1      & 94.2$\pm$0.1      & 55.1$\pm$0.6 & 91.9$\pm$0.9              \\
   \rowcolor{var2light}                      & KNN                     & 53.1$\pm$0.2      & 94.4$\pm$0.1      & 51.3$\pm$0.6       & 95.2$\pm$0.3        & 47.9$\pm$0.8      & 94.6$\pm$0.3      & 55.7$\pm$1.2 & 93.9$\pm$0.9              \\
   \rowcolor{var2light}                      & VIM                     & 53.5$\pm$0.2      & 94.7$\pm$0.2      & 51.8$\pm$0.8       & 95.6$\pm$0.2        & 47.4$\pm$0.7      & 95.0$\pm$0.1      & 59.9$\pm$0.2 & 92.0$\pm$1.2              \\
    \rowcolor{var2light}                     & SHE                     & 47.6$\pm$0.5      & 95.7$\pm$0.2      & 45.6$\pm$0.6       & 96.5$\pm$0.5        & \textbf{51.7$\pm$0.1}      & \textbf{94.2$\pm$0.1}      & 61.6$\pm$1.0 & 92.6$\pm$1.4              \\
   \rowcolor{var2light}                      & KLM                     & 48.2$\pm$0.0      & 94.1$\pm$0.1      & 47.3$\pm$1.0       & 95.5$\pm$0.5        & 50.6$\pm$0.6      & 95.6$\pm$0.5      & 53.7$\pm$0.1 & 91.4$\pm$1.1              \\
   \rowcolor{var2light}                      & OpenMax                 & 53.7$\pm$0.5      & 94.3$\pm$0.1      & 53.4$\pm$1.6       & 95.0$\pm$0.5        & 46.9$\pm$0.6      & 95.8$\pm$0.3      & 55.2$\pm$0.3 & 92.4$\pm$0.9              \\
   \rowcolor{var2dark}                      & MSP                     & 53.7$\pm$0.5      & 93.8$\pm$0.2      & 53.4$\pm$1.6       & 95.0$\pm$0.5        & 46.9$\pm$0.7      & 95.9$\pm$0.2      & 55.2$\pm$0.7 & 91.9$\pm$1.1              \\
 \rowcolor{var2dark} ResNet                        & MLS                     & 53.7$\pm$0.4      & 94.1$\pm$0.3      & 53.6$\pm$1.3       & 95.0$\pm$0.6        & 46.8$\pm$0.6      & 96.0$\pm$0.3      & 55.2$\pm$0.4 & \textbf{91.3$\pm$1.1}              \\
   \rowcolor{var2dark}                      & TempScale               & 53.7$\pm$0.5      & 93.8$\pm$0.2      & 53.4$\pm$1.6       & 95.0$\pm$0.5        & 46.9$\pm$0.7      & 95.9$\pm$0.2      & 55.2$\pm$0.7 & 91.9$\pm$1.1              \\
    \rowcolor{var2dark}                     & ODIN                    & 53.7$\pm$0.5      & 93.8$\pm$0.2      & 53.4$\pm$1.5       & 95.0$\pm$0.4        & 46.9$\pm$0.7      & 95.8$\pm$0.2      & 55.2$\pm$0.7 & 91.9$\pm$1.1              \\
      \rowcolor{var2dark}                   & EBO                     & 53.7$\pm$0.4      & 94.3$\pm$0.1      & 53.6$\pm$1.3       & 94.9$\pm$0.7        & 46.8$\pm$0.5      & 96.1$\pm$0.3      & 55.1$\pm$0.4 & 91.5$\pm$0.9              \\
      \rowcolor{var2dark}                   & GRAM                    & 52.5$\pm$0.2      & 100.0$\pm$0.0     & \textbf{55.6$\pm$1.1}       & 96.3$\pm$2.4        & 50.0$\pm$0.0      & 100.0$\pm$0.0     & 42.6$\pm$2.7 & 100.0$\pm$0.0             \\
      \rowcolor{var2dark}                   & GradNorm                & \textbf{53.8$\pm$0.7}      & \textbf{93.8$\pm$0.2}      & 55.0$\pm$1.2       & \textbf{94.5$\pm$0.5}        & 47.0$\pm$0.7      & 95.8$\pm$0.2      & 46.6$\pm$1.7 & 91.4$\pm$0.9              \\
       \rowcolor{var2dark}                  & ReAct                   & 53.3$\pm$0.3      & 94.3$\pm$0.1      & 52.6$\pm$1.4       & 95.4$\pm$0.6        & 47.0$\pm$0.8      & 95.9$\pm$0.2      & 58.0$\pm$0.3 & 92.0$\pm$1.1              \\
       \rowcolor{var2dark}                  & DICE                    & 52.9$\pm$0.2      & 94.2$\pm$0.4      & 52.6$\pm$1.9       & 94.8$\pm$0.3        & 47.3$\pm$0.6      & 95.2$\pm$0.4      & 51.8$\pm$1.4 & 92.9$\pm$0.2              \\
        \rowcolor{var2dark}                 & ASH                     & 53.5$\pm$0.6      & 94.3$\pm$0.4      & 52.8$\pm$1.2       & 95.1$\pm$0.6        & 46.8$\pm$0.5      & 96.1$\pm$0.4      & 53.9$\pm$0.5 & 91.8$\pm$1.5              \\
\midrule
\rowcolor{var3light} \multirow{17}{*}{FT-T}   & MDS                     & 51.2$\pm$0.2      & 95.2$\pm$0.3      & 52.0$\pm$0.7       & 95.7$\pm$0.1        & 51.6$\pm$0.4      & \textbf{94.2$\pm$0.1}      & 61.9$\pm$0.7 & \textbf{90.1$\pm$1.1}              \\
  \rowcolor{var3light}                        & RMDS                    & 51.6$\pm$0.8      & 94.0$\pm$0.1      & 50.8$\pm$0.5       & 94.5$\pm$1.0        & 49.2$\pm$0.2      & 95.6$\pm$0.2      & 53.8$\pm$1.0 & 90.3$\pm$1.5              \\
   \rowcolor{var3light}                       & KNN                     & 52.7$\pm$0.3      & 95.4$\pm$0.2      & 53.1$\pm$0.3       & 95.3$\pm$0.4        & 48.8$\pm$0.2      & 95.0$\pm$0.5      & \textbf{63.3$\pm$1.1} & 92.4$\pm$1.6              \\
   \rowcolor{var3light}                       & VIM                     & 53.7$\pm$0.3      & 95.1$\pm$0.3      & 53.7$\pm$0.1       & 95.1$\pm$0.2        & 47.3$\pm$0.2      & 95.4$\pm$0.2      & 62.5$\pm$0.3 & 89.6$\pm$1.1              \\
   \rowcolor{var3light}                       & SHE                     & 49.5$\pm$0.2      & 94.4$\pm$0.4      & 50.5$\pm$0.8       & 94.6$\pm$0.2        & 51.3$\pm$0.2      & 95.2$\pm$0.4      & 55.4$\pm$0.6 & 90.5$\pm$1.4              \\
   \rowcolor{var3light}                       & KLM                     & 47.6$\pm$0.6      & 93.9$\pm$0.3      & 46.0$\pm$0.9       & \textbf{94.1$\pm$0.3}        & \textbf{52.1$\pm$0.5}      & 95.5$\pm$0.5      & 46.1$\pm$1.0 & 91.2$\pm$1.0              \\
    \rowcolor{var3light}                      & OpenMax                 & \textbf{55.1$\pm$0.5}      & \textbf{93.8$\pm$0.1}      & 55.4$\pm$0.1       & 94.3$\pm$0.3        & 46.3$\pm$0.4      & 95.5$\pm$0.1      & 62.5$\pm$0.1 & 90.2$\pm$1.0              \\
    \rowcolor{var3dark}                      & MSP                     & 55.1$\pm$0.5      & 93.8$\pm$0.1      & 55.4$\pm$0.1       & 94.3$\pm$0.3        & 46.4$\pm$0.4      & 95.4$\pm$0.3      & 62.6$\pm$0.1 & 90.2$\pm$1.0              \\
\rowcolor{var3dark}  FT-T                         & MLS                     & 55.0$\pm$0.5      & 94.0$\pm$0.1      & 55.4$\pm$0.1       & 94.2$\pm$0.4        & 46.4$\pm$0.3      & 95.6$\pm$0.1      & 62.2$\pm$0.1 & 90.4$\pm$0.9              \\
   \rowcolor{var3dark}                       & TempScale               & 55.1$\pm$0.5      & 93.8$\pm$0.1      & 55.4$\pm$0.1       & 94.3$\pm$0.3        & 46.4$\pm$0.4      & 95.4$\pm$0.3      & 62.6$\pm$0.1 & 90.2$\pm$1.0              \\
   \rowcolor{var3dark}                       & ODIN                    & 55.1$\pm$0.5      & 93.8$\pm$0.1      & 55.4$\pm$0.1       & 94.3$\pm$0.3        & 46.4$\pm$0.4      & 95.4$\pm$0.3      & 62.6$\pm$0.1 & 90.3$\pm$1.0              \\
    \rowcolor{var3dark}                      & EBO                     & 55.0$\pm$0.5      & 94.0$\pm$0.1      & 55.4$\pm$0.1       & 94.6$\pm$0.3        & 46.4$\pm$0.3      & 95.6$\pm$0.3      & 62.2$\pm$0.2 & 90.9$\pm$1.2              \\
   \rowcolor{var3dark}                       & GRAM                    & 54.3$\pm$1.2      & 94.4$\pm$0.2      & 54.7$\pm$0.8       & 94.6$\pm$0.2        & 46.6$\pm$0.9      & 95.7$\pm$0.3      & 55.8$\pm$5.4 & 89.9$\pm$1.5              \\
      \rowcolor{var3dark}                    & GradNorm                & 53.5$\pm$0.1      & 93.8$\pm$0.1      & 51.1$\pm$4.3       & 94.6$\pm$0.4        & 47.2$\pm$1.3      & 95.4$\pm$0.3      & 49.3$\pm$5.8 & 90.5$\pm$1.1              \\
    \rowcolor{var3dark}                      & ReAct                   & 54.9$\pm$0.4      & 94.0$\pm$0.1      & \textbf{55.6$\pm$0.1}       & 94.3$\pm$0.3        & 46.4$\pm$0.4      & 95.4$\pm$0.4      & 61.7$\pm$0.8 & 90.9$\pm$1.4              \\
      \rowcolor{var3dark}                    & DICE                    & 54.9$\pm$0.5      & 94.3$\pm$0.5      & 55.3$\pm$0.1       & 94.4$\pm$0.2        & 46.6$\pm$0.5      & 95.5$\pm$0.2      & 61.6$\pm$0.2 & 90.4$\pm$0.9              \\
       \rowcolor{var3dark}                   & ASH                     & 54.7$\pm$0.6      & 94.3$\pm$0.2      & 51.4$\pm$1.6       & 95.1$\pm$0.5        & 46.7$\pm$0.2      & 95.4$\pm$0.3      & 54.8$\pm$2.2 & 90.5$\pm$1.4  \\
                         \bottomrule
\end{tabular}
\caption{
Comparing OOD detection methods in the \textit{near-OOD} setting, with performance evaluated through AUROC and FPR@95.
In eICU, distinguishing variables are gender (females as ID) and ethnicity (African Americans as ID). In MIMIC-IV, distinguishing variables are gender (females as ID) and admission type (surgical on the same day of admission as ID). Results are averaged over 3 runs. For each model, the best value in each column is shown in bold, and distance-based post-hoc approaches are distinguished with a lighter color.}
\label{tab:other_near}
\end{table*}

\section{FPR@95 Results}\label{apd:fpr}

In Tables \ref{tab:near_far_ood} and \ref{tab:synthesized_ood}, we presented the results of our analyses, highlighting the performance based on the AUROC. While AUROC serves as a measure of the discriminative capability between novelty scores assigned to ID and OOD data, devoid of any reliance on a predetermined threshold, it would be beneficial to complement this assessment by evaluating the performance at a specific threshold. 
For this purpose, we have measured the false positive rate at the $\lambda$ corresponding to the TPR equal to 95\% as an acceptable rate for TPR, which is known as FPR@95. 

By replacing AUROC with FPR@95 in Tables \ref{tab:near_far_ood} and \ref{tab:synthesized_ood}, we have replicated the corresponding outcomes within Tables \ref{tab:near_far_ood_fpr} and \ref{tab:synthesized_ood_fpr}. 
These new tables are in agreement with our conclusions, and if one method has a relatively high AUROC, it would mostly have a relatively low FPR@95. However, it is helpful for comparing methods at a specific threshold. For example, RMDS and KLM applied to an MLP model are achieving similar AUROCs, but RMDS is definitely better at the measured threshold.

% \usepackage{multirow}
\begin{table*}[]
\footnotesize
%\scriptsize
\centering
\renewcommand{\arraystretch}{0.85}
\setlength{\tabcolsep}{4pt}
\begin{tabular}{c|c|ccc|ccc}
\toprule
\multirow{2}{*}{Model}   & \multirow{2}{*}{Method} & \multicolumn{3}{c|}{eICU}                        & \multicolumn{3}{c}{MIMIC-IV}                    \\
                         &                         & Far-OOD       & Near-OOD (Eth) & Near-OOD (Age) & Far-OOD       & Near-OOD (FCU) & Near-OOD (Age) \\ \midrule
     \rowcolor{var4}                     & AE                      & 17.5$\pm$5.3  & 93.3$\pm$1.6   & 94.3$\pm$0.2   & 0.2$\pm$0.3   & 72.7$\pm$11.5  & \textbf{90.7$\pm$0.7}   \\
    \rowcolor{var4}                     & VAE                     & 22.5$\pm$7.4  & 93.3$\pm$1.5   & 94.6$\pm$0.2   & \textbf{0.2$\pm$0.3}   & 71.9$\pm$11.8  & 90.7$\pm$0.8   \\
\rowcolor{var4}  & HI-VAE                    & 83.4$\pm$1.5 &  100.0$\pm$0.0  & 100.0$\pm$0.0 & 99.9$\pm$0.1 & \textbf{44.5$\pm$21.9}  & 90.9$\pm$6.9 \\
\rowcolor{var4} Density                  & Flow                    & \textbf{0.0$\pm$0.0}   & \textbf{89.6$\pm$4.9}   & 94.7$\pm$0.4   & 35.9$\pm$39.2 & 96.8$\pm$1.1   & 93.5$\pm$1.0   \\
\rowcolor{var4} Based                    & PPCA                    & 15.5$\pm$4.3  & 91.5$\pm$3.5   & \textbf{93.8$\pm$0.3}   & 0.3$\pm$0.3   & 76.7$\pm$9.8   & 92.0$\pm$0.6   \\
\rowcolor{var4}                         & LOF                     & 17.4$\pm$2.3  & 92.8$\pm$2.4   & 95.5$\pm$0.2   & 3.5$\pm$0.6   & 78.4$\pm$5.4   & 91.9$\pm$0.4   \\
 \rowcolor{var4}                        & DUE                     & 80.9$\pm$6.1  & 94.3$\pm$1.6   & 94.5$\pm$0.1   & 7.6$\pm$2.5   & 88.6$\pm$3.2   & 91.1$\pm$0.4   \\
\midrule
\rowcolor{var1light} \multirow{17}{*}{MLP}    & MDS                     & \textbf{62.7$\pm$6.6}  & \textbf{91.1$\pm$2.6}  & 94.6$\pm$0.3   & \textbf{4.3$\pm$3.2}   & 85.2$\pm$2.2   & \textbf{92.5$\pm$0.6}   \\
\rowcolor{var1light}                         & RMDS                    & 90.3$\pm$5.0  & 94.7$\pm$1.5   & 95.7$\pm$0.2   & 60.0$\pm$29.2 & 89.8$\pm$5.3   & 95.0$\pm$0.1   \\
 \rowcolor{var1light}                        & KNN                     & 73.3$\pm$13.1 & 93.5$\pm$3.7   & 95.9$\pm$0.2   & 81.9$\pm$18.8 & \textbf{78.8$\pm$3.8}   & 93.4$\pm$0.6   \\
\rowcolor{var1light}                         & VIM                     & 84.9$\pm$7.3  & 94.2$\pm$1.8   & 95.4$\pm$0.5   & 9.5$\pm$6.4   & 82.3$\pm$1.1   & 93.0$\pm$0.8   \\
  \rowcolor{var1light}                       & SHE                     & 87.8$\pm$1.1  & 94.4$\pm$1.4   & \textbf{94.0$\pm$0.8}   & 26.6$\pm$15.7 & 93.3$\pm$1.5   & 96.1$\pm$0.4   \\
  \rowcolor{var1light}                       & KLM                     & 93.3$\pm$1.1  & 94.7$\pm$1.5   & 95.8$\pm$0.2   & 82.2$\pm$16.5 & 90.8$\pm$5.1   & 94.2$\pm$0.5   \\
  \rowcolor{var1light}                       & OpenMax                 & 93.8$\pm$3.9  & 95.9$\pm$1.4   & 95.9$\pm$0.1   & 70.4$\pm$24.4 & 87.0$\pm$2.1   & 94.7$\pm$0.9   \\
  \rowcolor{var1dark}                       & MSP                     & 94.7$\pm$3.4  & 97.2$\pm$2.0   & 95.9$\pm$0.2   & 99.1$\pm$1.0  & 91.6$\pm$6.5   & 94.7$\pm$0.7   \\
\rowcolor{var1dark} MLP                         & MLS                     & 94.6$\pm$3.6  & 95.4$\pm$1.3   & 96.0$\pm$0.2   & 98.3$\pm$1.6  & 87.8$\pm$3.9   & 94.5$\pm$0.6   \\
\rowcolor{var1dark}                         & TempScale               & 94.7$\pm$3.4  & 95.9$\pm$1.1   & 95.9$\pm$0.2   & 99.1$\pm$1.0  & 89.1$\pm$6.4   & 94.7$\pm$0.7   \\
 \rowcolor{var1dark}                        & ODIN                    & 94.7$\pm$3.4  & 95.4$\pm$1.3   & 95.9$\pm$0.3   & 98.9$\pm$1.1  & 88.0$\pm$4.0   & 94.6$\pm$0.8   \\
  \rowcolor{var1dark}                       & EBO                     & 94.5$\pm$3.7  & 95.4$\pm$1.3   & 96.0$\pm$0.2   & 96.8$\pm$3.1  & 87.8$\pm$3.9   & 94.8$\pm$0.6   \\
   \rowcolor{var1dark}                      & GRAM                    & 99.4$\pm$0.3  & 98.1$\pm$3.0   & 100.0$\pm$0.0  & 100.0$\pm$0.0 & 98.7$\pm$2.1   & 100.0$\pm$0.0  \\
    \rowcolor{var1dark}                     & GradNorm                & 95.3$\pm$2.7  & 95.5$\pm$1.2   & 95.8$\pm$0.2   & 99.1$\pm$0.8  & 88.4$\pm$3.9   & 94.6$\pm$0.6   \\
   \rowcolor{var1dark}                      & ReAct                   & 93.6$\pm$2.4  & 95.4$\pm$1.0   & 96.0$\pm$0.2   & 72.5$\pm$17.4 & 87.7$\pm$3.8   & 94.8$\pm$0.6   \\
   \rowcolor{var1dark}                      & DICE                    & 95.8$\pm$2.5  & 95.3$\pm$0.9   & 95.0$\pm$0.2   & 97.8$\pm$3.2  & 88.2$\pm$3.8   & 94.2$\pm$1.1   \\
   \rowcolor{var1dark}                      & ASH                     & 94.3$\pm$3.7  & 95.9$\pm$0.9   & 95.8$\pm$0.4   & 97.2$\pm$2.7  & 88.7$\pm$2.7   & 94.7$\pm$0.3   \\
\midrule
\rowcolor{var2light} \multirow{17}{*}{ResNet} & MDS                     & \textbf{14.0$\pm$1.9}  & 90.5$\pm$3.3   & \textbf{94.0$\pm$0.2}   & \textbf{0.4$\pm$0.4}   & \textbf{77.4$\pm$9.2}   & 92.1$\pm$0.9   \\
\rowcolor{var2light}                          & RMDS                    & 93.0$\pm$4.1  & 95.0$\pm$1.8   & 95.1$\pm$0.1   & 54.9$\pm$32.1 & 88.2$\pm$8.1   & 94.7$\pm$0.4   \\
 \rowcolor{var2light}                         & KNN                     & 39.4$\pm$7.0  & \textbf{88.3$\pm$3.3}   & 95.3$\pm$0.1   & 32.9$\pm$17.0 & 93.0$\pm$2.2   & 93.4$\pm$0.6   \\
 \rowcolor{var2light}                         & VIM                     & 28.7$\pm$2.2  & 89.8$\pm$2.0   & 94.5$\pm$0.2   & 1.1$\pm$1.1   & 81.3$\pm$2.2   & \textbf{91.8$\pm$0.8}   \\
 \rowcolor{var2light}                         & SHE                     & 54.2$\pm$4.7  & 94.8$\pm$0.6   & 94.2$\pm$0.2   & 0.5$\pm$0.6   & 79.2$\pm$7.2   & 93.3$\pm$0.4   \\
 \rowcolor{var2light}                         & KLM                     & 90.9$\pm$3.2  & 95.0$\pm$0.7   & 95.1$\pm$0.3   & 88.5$\pm$8.6  & 89.0$\pm$5.7   & 93.9$\pm$0.1   \\
    \rowcolor{var2light}                      & OpenMax                 & 85.2$\pm$4.6  & 95.8$\pm$1.5   & 95.7$\pm$0.6   & 62.6$\pm$31.8 & 88.8$\pm$6.4   & 93.6$\pm$0.2   \\
   \rowcolor{var2dark}                       & MSP                     & 89.3$\pm$2.9  & 94.5$\pm$0.9   & 95.5$\pm$0.2   & 97.1$\pm$3.5  & 89.6$\pm$7.1   & 94.7$\pm$0.3   \\
\rowcolor{var2dark}   ResNet                         & MLS                     & 86.2$\pm$4.4  & 94.3$\pm$1.2   & 95.7$\pm$0.5   & 88.3$\pm$8.6  & 88.5$\pm$7.4   & 94.2$\pm$0.1   \\
 \rowcolor{var2dark}                          & TempScale               & 89.3$\pm$2.9  & 94.5$\pm$0.9   & 95.5$\pm$0.2   & 97.1$\pm$3.5  & 89.6$\pm$7.1   & 94.7$\pm$0.3   \\
 \rowcolor{var2dark}                          & ODIN                    & 89.3$\pm$2.9  & 94.5$\pm$0.9   & 95.5$\pm$0.2   & 97.0$\pm$3.5  & 89.6$\pm$7.0   & 94.8$\pm$0.3   \\
   \rowcolor{var2dark}                        & EBO                     & 84.8$\pm$5.5  & 94.3$\pm$1.2   & 95.7$\pm$0.8   & 86.9$\pm$10.1 & 87.7$\pm$8.9   & 93.3$\pm$0.6   \\
     \rowcolor{var2dark}                      & GRAM                    & 99.8$\pm$0.3  & 95.2$\pm$1.2   & 100.0$\pm$0.0  & 100.0$\pm$0.0 & 99.5$\pm$0.4   & 100.0$\pm$0.0  \\
    \rowcolor{var2dark}                       & GradNorm                & 94.7$\pm$2.4  & 94.7$\pm$1.7   & 95.6$\pm$0.2   & 99.1$\pm$1.1  & 91.5$\pm$6.1   & 94.8$\pm$0.2   \\
    \rowcolor{var2dark}                       & ReAct                   & 79.7$\pm$6.8  & 94.0$\pm$1.4   & 95.5$\pm$0.6   & 60.1$\pm$32.3 & 81.2$\pm$4.8   & 93.0$\pm$0.5   \\
 \rowcolor{var2dark}                          & DICE                    & 95.2$\pm$1.4  & 95.9$\pm$0.9   & 95.4$\pm$0.2   & 98.6$\pm$2.6  & 90.9$\pm$6.2   & 93.3$\pm$0.6   \\
    \rowcolor{var2dark}                       & ASH                     & 81.2$\pm$5.9  & 93.3$\pm$2.8   & 95.7$\pm$0.6   & 86.5$\pm$17.5 & 88.9$\pm$7.0   & 93.4$\pm$0.7   \\
\midrule
\rowcolor{var3light}   \multirow{17}{*}{FT-T}   & MDS                     & 61.1$\pm$14.6 & \textbf{89.7$\pm$4.3}   & 94.3$\pm$0.5   & 29.5$\pm$11.5 & 73.8$\pm$7.9   & 94.2$\pm$0.9   \\
   \rowcolor{var3light}                      & RMDS                    & 91.1$\pm$4.9  & 93.1$\pm$1.6   & 95.6$\pm$0.1   & 64.4$\pm$25.9 & 88.5$\pm$6.8   & 94.4$\pm$0.3   \\
   \rowcolor{var3light}                      & KNN                     & \textbf{53.2$\pm$13.4} & 91.8$\pm$2.0   & 94.7$\pm$0.2   & 38.4$\pm$22.3 & \textbf{70.7$\pm$10.5}  & 93.9$\pm$0.1   \\
    \rowcolor{var3light}                     & VIM                     & 57.6$\pm$11.5 & 90.7$\pm$4.3   & \textbf{94.6$\pm$0.0}   & \textbf{18.1$\pm$2.1}  & 76.1$\pm$1.6   & \textbf{92.5$\pm$0.6}   \\
     \rowcolor{var3light}                    & SHE                     & 84.7$\pm$2.6  & 93.2$\pm$1.8   & 95.1$\pm$0.3   & 52.5$\pm$17.1 & 83.8$\pm$5.9   & 96.7$\pm$0.2   \\
    \rowcolor{var3light}                     & KLM                     & 89.5$\pm$5.1  & 92.8$\pm$1.4   & 95.2$\pm$0.3   & 70.4$\pm$8.8  & 86.8$\pm$9.0   & 95.7$\pm$0.5   \\
    \rowcolor{var3light}                     & OpenMax                 & 80.4$\pm$1.9  & 93.8$\pm$1.9   & 95.2$\pm$0.1   & 78.6$\pm$17.3 & 78.3$\pm$3.7   & 94.3$\pm$0.3   \\
    \rowcolor{var3dark}                     & MSP                     & 80.4$\pm$2.0  & 93.5$\pm$1.6   & 95.3$\pm$0.2   & 80.2$\pm$15.2 & 77.8$\pm$4.4   & 95.0$\pm$0.2   \\
\rowcolor{var3dark}  FT-T                           & MLS                     & 80.7$\pm$2.0  & 93.4$\pm$1.6   & 95.3$\pm$0.1   & 82.5$\pm$10.9 & 77.3$\pm$3.9   & 95.1$\pm$0.2   \\
\rowcolor{var3dark}                        & TempScale               & 80.4$\pm$2.0  & 93.5$\pm$1.6   & 95.3$\pm$0.2   & 80.2$\pm$15.2 & 77.8$\pm$4.4   & 95.0$\pm$0.2   \\
  \rowcolor{var3dark}                        & ODIN                    & 80.3$\pm$2.0  & 93.5$\pm$1.6   & 95.3$\pm$0.2   & 80.1$\pm$15.2 & 77.8$\pm$4.2   & 95.0$\pm$0.2   \\
   \rowcolor{var3dark}                       & EBO                     & 80.9$\pm$2.1  & 93.3$\pm$1.6   & 95.2$\pm$0.1   & 83.1$\pm$9.5  & 78.1$\pm$3.2   & 94.6$\pm$0.3   \\
    \rowcolor{var3dark}                      & GRAM                    & 87.7$\pm$2.7  & 93.6$\pm$2.3   & 95.0$\pm$0.3   & 67.6$\pm$18.8 & 77.6$\pm$7.8   & 94.5$\pm$0.3   \\
    \rowcolor{var3dark}                      & GradNorm                & 87.2$\pm$2.1  & 92.0$\pm$3.2   & 95.3$\pm$0.2   & 78.1$\pm$14.4 & 86.0$\pm$6.5   & 95.0$\pm$0.2   \\
     \rowcolor{var3dark}                     & ReAct                   & 81.0$\pm$2.0  & 94.5$\pm$2.3   & 95.2$\pm$0.2   & 82.4$\pm$9.2  & 80.1$\pm$1.3   & 94.2$\pm$0.3   \\
       \rowcolor{var3dark}                   & DICE                    & 80.1$\pm$2.1  & 93.0$\pm$2.2   & 94.9$\pm$0.1   & 57.8$\pm$7.8  & 80.2$\pm$4.4   & 92.4$\pm$0.8   \\
        \rowcolor{var3dark}                  & ASH                     & 81.2$\pm$2.6  & 92.6$\pm$2.1   & 95.3$\pm$0.1   & 92.1$\pm$8.0  & 80.5$\pm$5.5   & 95.3$\pm$0.6  \\ \bottomrule
\end{tabular}
\caption{
Comparing OOD detection methods in the \textit{far-OOD} and \textit{near-OOD} settings using eICU and MIMIC-IV datasets, with performance evaluated through FPR@95. This Table is similar to Table \ref{tab:near_far_ood}, but assesses FPR@95 instead of utilizing AUROC. 
%The name in the top row shows the dataset considered as ID in the far-OOD experiment and divided for the near-OOD test.
FCU and Eth stand for first care unit and ethnicity as the variables utilized for constructing the ID/OOD sets.
Results are averaged over 5 runs.
For each model, the best value in each column is shown in bold, and distance-based post-hoc approaches are distinguished with a lighter color.
}
\label{tab:near_far_ood_fpr}
\end{table*}

% \usepackage{multirow}
\begin{table*}[]
\footnotesize
%\scriptsize
\centering
\renewcommand{\arraystretch}{0.85}
\setlength{\tabcolsep}{10pt}
\begin{tabular}{c|c|ccc|ccc}
\toprule
\multirow{2}{*}{Model}   & \multirow{2}{*}{Method} & \multicolumn{3}{c|}{eICU}                                  & \multicolumn{3}{c}{MIMIC-IV}                              \\ 
                         &                         & $\mathcal{F}$=10 & $\mathcal{F}$=100 & $\mathcal{F}$=1000 & $\mathcal{F}$=10 & $\mathcal{F}$=100 & $\mathcal{F}$=1000 \\ \midrule
\rowcolor{var4}                          & AE                      & 47.1$\pm$3.9     & 25.4$\pm$3.7      & 22.0$\pm$4.1       & 51.6$\pm$3.8     & 35.3$\pm$3.2      & 28.5$\pm$4.2       \\
 \rowcolor{var4}                        & VAE                     & 47.9$\pm$4.0     & 25.6$\pm$3.7      & 22.1$\pm$4.2       & 51.5$\pm$3.7     & 35.1$\pm$3.1      & 28.4$\pm$4.1       \\
\rowcolor{var4}  & HI-VAE                    & 94.9$\pm$0.2 & 94.7$\pm$0.2 & 94.7$\pm$0.5  & 94.9$\pm$0.4 & 94.5$\pm$0.6  & 94.1$\pm$0.9 \\
\rowcolor{var4} Density                  & Flow                    & 60.2$\pm$7.2     & 36.8$\pm$7.4      & 26.5$\pm$5.2       & 82.1$\pm$3.1     & 62.6$\pm$3.3      & 44.6$\pm$2.1       \\
\rowcolor{var4} Based                    & PPCA                    & 46.6$\pm$2.7     & 25.8$\pm$4.4      & 22.8$\pm$5.0       & 50.7$\pm$3.3     & 35.0$\pm$3.1      & 28.6$\pm$4.2       \\
 \rowcolor{var4}                        & LOF                     & \textbf{35.4$\pm$3.6}     & \textbf{23.9$\pm$4.3}      & \textbf{21.5$\pm$4.4}       & \textbf{46.4$\pm$3.2}     & \textbf{32.6$\pm$3.7}      & \textbf{27.5$\pm$3.5}       \\
\rowcolor{var4}                        & DUE                     & 78.2$\pm$4.6     & 47.9$\pm$2.5      & 25.8$\pm$3.9       & 83.1$\pm$2.5     & 53.0$\pm$4.7      & 37.2$\pm$2.7       \\
\midrule
\rowcolor{var1light} \multirow{17}{*}{MLP}    & MDS                     & \textbf{67.9$\pm$5.4}     & \textbf{42.3$\pm$3.2}      & \textbf{24.5$\pm$4.0}       & \textbf{67.1$\pm$4.7}     & \textbf{43.0$\pm$5.1}      & \textbf{32.6$\pm$3.6}       \\
 \rowcolor{var1light}                         & RMDS                    & 77.5$\pm$0.9     & 50.6$\pm$2.6      & 30.7$\pm$4.0       & 87.9$\pm$2.0     & 66.2$\pm$2.4      & 45.4$\pm$2.1       \\
   \rowcolor{var1light}                       & KNN                     & 82.8$\pm$8.4     & 73.1$\pm$12.8     & 67.9$\pm$15.8      & 79.9$\pm$6.1     & 61.7$\pm$14.6     & 53.0$\pm$18.5      \\
   \rowcolor{var1light}                       & VIM                     & 86.4$\pm$2.2     & 77.1$\pm$2.6      & 69.0$\pm$5.1       & 79.0$\pm$9.1     & 59.8$\pm$16.8     & 51.0$\pm$17.6      \\
     \rowcolor{var1light}                     & SHE                     & 77.1$\pm$4.7     & 48.0$\pm$2.8      & 24.5$\pm$4.0       & 77.5$\pm$1.7     & 48.4$\pm$6.2      & 35.0$\pm$4.1       \\
      \rowcolor{var1light}                    & KLM                     & 86.3$\pm$8.5     & 77.5$\pm$17.9     & 71.2$\pm$22.7      & 88.6$\pm$4.8     & 74.5$\pm$11.6     & 65.5$\pm$15.9      \\
     \rowcolor{var1light}                     & OpenMax                 & 83.9$\pm$0.8     & 60.6$\pm$2.4      & 42.1$\pm$4.1       & 91.3$\pm$1.1     & 74.5$\pm$3.1      & 55.1$\pm$6.7       \\
   \rowcolor{var1dark}                       & MSP                     & 94.0$\pm$0.4     & 95.6$\pm$0.6      & 98.4$\pm$1.0       & 93.7$\pm$0.6     & 96.0$\pm$0.6      & 97.1$\pm$0.7       \\
    \rowcolor{var1dark} MLP                    & MLS                     & 94.1$\pm$0.4     & 95.8$\pm$0.9      & 98.5$\pm$0.9       & 93.4$\pm$0.8     & 93.8$\pm$2.1      & 93.2$\pm$3.1       \\
    \rowcolor{var1dark}                     & TempScale               & 94.0$\pm$0.4     & 95.6$\pm$0.6      & 98.4$\pm$0.9       & 93.8$\pm$0.6     & 96.1$\pm$0.6      & 97.2$\pm$0.6       \\
    \rowcolor{var1dark}                     & ODIN                    & 94.0$\pm$0.4     & 95.6$\pm$0.6      & 98.4$\pm$1.0       & 93.7$\pm$0.6     & 96.0$\pm$0.5      & 97.1$\pm$0.6       \\
    \rowcolor{var1dark}                     & EBO                     & 94.2$\pm$0.4     & 95.7$\pm$1.1      & 98.5$\pm$0.9       & 93.3$\pm$1.1     & 93.4$\pm$2.5      & 92.9$\pm$3.3       \\
     \rowcolor{var1dark}                    & GRAM                    & 98.0$\pm$1.7     & 98.7$\pm$0.8      & 99.9$\pm$0.0       & 99.9$\pm$0.1     & 99.9$\pm$0.1      & 99.9$\pm$0.1       \\
    \rowcolor{var1dark}                     & GradNorm                & 94.2$\pm$0.5     & 96.0$\pm$0.4      & 98.9$\pm$0.7       & 94.3$\pm$0.5     & 96.9$\pm$0.4      & 97.7$\pm$0.6       \\
    \rowcolor{var1dark}                     & ReAct                   & 92.4$\pm$2.3     & 91.8$\pm$4.0      & 91.4$\pm$5.3       & 92.5$\pm$1.2     & 89.7$\pm$3.0      & 86.8$\pm$4.6       \\
    \rowcolor{var1dark}                     & DICE                    & 93.1$\pm$0.6     & 93.7$\pm$0.9      & 95.5$\pm$2.0       & 93.2$\pm$0.4     & 94.6$\pm$0.9      & 96.0$\pm$2.0       \\
    \rowcolor{var1dark}                     & ASH                     & 94.2$\pm$0.5     & 95.6$\pm$0.8      & 98.1$\pm$1.0       & 93.6$\pm$0.8     & 93.4$\pm$2.3      & 92.6$\pm$3.0       \\
\midrule
\rowcolor{var2light} \multirow{17}{*}{ResNet} & MDS                     & \textbf{57.7$\pm$4.5}     & \textbf{26.6$\pm$4.5}      & 20.6$\pm$4.9      & \textbf{60.9$\pm$2.3}     & \textbf{39.7$\pm$2.7}      & \textbf{30.7$\pm$3.0}       \\
 \rowcolor{var2light}                        & RMDS                    & 87.2$\pm$2.7     & 64.5$\pm$6.3      & 39.1$\pm$6.0       & 91.6$\pm$1.7     & 70.7$\pm$2.9      & 47.9$\pm$4.0       \\
  \rowcolor{var2light}                       & KNN                     & 69.4$\pm$4.9     & 41.4$\pm$6.1      & 33.0$\pm$6.5       & 70.3$\pm$3.3     & 46.7$\pm$0.7      & 35.5$\pm$1.8       \\
  \rowcolor{var2light}                       & VIM                     & 60.9$\pm$4.8     & 26.9$\pm$4.1      & \textbf{19.7$\pm$4.3}       & 65.8$\pm$0.8     & 42.4$\pm$3.5      & 31.9$\pm$3.7       \\
 \rowcolor{var2light}                        & SHE                     & 68.3$\pm$3.2     & 34.4$\pm$2.8      & 20.4$\pm$3.0       & 71.4$\pm$4.6     & 44.2$\pm$2.9      & 32.3$\pm$2.7       \\
  \rowcolor{var2light}                       & KLM                     & 88.4$\pm$1.3     & 72.7$\pm$8.0      & 58.0$\pm$14.1      & 90.9$\pm$1.3     & 74.8$\pm$4.9      & 63.7$\pm$8.5       \\
   \rowcolor{var2light}                      & OpenMax                 & 83.0$\pm$1.0     & 64.0$\pm$6.0      & 42.9$\pm$9.6       & 87.2$\pm$3.7     & 69.7$\pm$2.1      & 61.0$\pm$13.3      \\
    \rowcolor{var2dark}                     & MSP                     & 91.4$\pm$2.5     & 93.6$\pm$1.6      & 97.7$\pm$0.4       & 90.6$\pm$3.1     & 94.1$\pm$1.7      & 97.5$\pm$0.2       \\
   \rowcolor{var2dark}           ResNet            & MLS                     & 90.8$\pm$1.6     & 92.0$\pm$1.3      & 94.6$\pm$2.2       & 89.9$\pm$3.6     & 89.8$\pm$5.4      & 89.1$\pm$7.6       \\
  \rowcolor{var2dark}                        & TempScale               & 91.4$\pm$2.5     & 93.6$\pm$1.6      & 97.8$\pm$0.4       & 90.7$\pm$3.1     & 94.1$\pm$1.7      & 97.5$\pm$0.2       \\
  \rowcolor{var2dark}                        & ODIN                    & 91.4$\pm$2.5     & 93.6$\pm$1.6      & 97.7$\pm$0.4       & 90.5$\pm$3.0     & 94.1$\pm$1.6      & 97.5$\pm$0.3       \\
   \rowcolor{var2dark}                       & EBO                     & 90.6$\pm$1.5     & 91.8$\pm$1.5      & 94.4$\pm$2.3       & 89.7$\pm$3.8     & 89.0$\pm$6.1      & 88.8$\pm$8.2       \\
   \rowcolor{var2dark}                       & GRAM                    & 98.6$\pm$1.6     & 99.9$\pm$0.1      & 100.0$\pm$0.0      & 99.8$\pm$0.2     & 99.9$\pm$0.1      & 99.9$\pm$0.1       \\
    \rowcolor{var2dark}                      & GradNorm                & 95.0$\pm$0.4     & 98.2$\pm$1.0      & 99.4$\pm$0.4       & 94.2$\pm$0.4     & 97.5$\pm$0.7      & 98.4$\pm$0.2       \\
    \rowcolor{var2dark}                      & ReAct                   & 83.7$\pm$5.9     & 71.9$\pm$14.7     & 64.8$\pm$19.4      & 86.1$\pm$7.3     & 75.7$\pm$15.0     & 70.2$\pm$19.0      \\
     \rowcolor{var2dark}                     & DICE                    & 94.3$\pm$1.4     & 97.4$\pm$0.5      & 99.2$\pm$0.6       & 90.4$\pm$3.7     & 92.6$\pm$2.6      & 95.6$\pm$1.2       \\
      \rowcolor{var2dark}                    & ASH                     & 91.1$\pm$1.0     & 91.5$\pm$1.9      & 93.7$\pm$3.1       & 90.3$\pm$3.4     & 89.7$\pm$5.2      & 89.1$\pm$7.6       \\
\midrule
\rowcolor{var3light}  \multirow{17}{*}{FT-T}   & MDS                     & \textbf{81.6$\pm$1.3}     & \textbf{48.1$\pm$4.1}      & 28.0$\pm$4.9       & \textbf{82.3$\pm$3.6}     & \textbf{52.5$\pm$2.4}      & \textbf{38.6$\pm$4.7}       \\
  \rowcolor{var3light}                        & RMDS                    & 91.9$\pm$2.3     & 78.7$\pm$10.4     & 64.2$\pm$17.7      & 94.1$\pm$0.3     & 79.0$\pm$8.2      & 58.2$\pm$11.6      \\
   \rowcolor{var3light}                       & KNN                     & 83.6$\pm$1.8     & 53.4$\pm$5.3      & 29.8$\pm$5.6       & 85.5$\pm$2.3     & 55.0$\pm$1.7      & 39.7$\pm$4.6       \\
    \rowcolor{var3light}                      & VIM                     & 81.9$\pm$2.8     & 48.7$\pm$3.5      & \textbf{26.6$\pm$3.0}       & 85.5$\pm$4.3     & 54.1$\pm$2.1      & 40.5$\pm$4.5       \\
     \rowcolor{var3light}                     & SHE                     & 88.2$\pm$2.5     & 63.0$\pm$6.6      & 41.6$\pm$4.4       & 91.2$\pm$1.3     & 68.4$\pm$2.9      & 43.3$\pm$5.3       \\
      \rowcolor{var3light}                    & KLM                     & 91.0$\pm$1.8     & 79.5$\pm$4.4      & 60.9$\pm$9.7       & 93.2$\pm$0.9     & 81.8$\pm$3.6      & 76.5$\pm$7.2       \\
      \rowcolor{var3light}                    & OpenMax                 & 91.2$\pm$1.9     & 80.7$\pm$4.8      & 60.6$\pm$9.2       & 94.2$\pm$0.5     & 83.9$\pm$4.8      & 75.0$\pm$13.1      \\
     \rowcolor{var3dark}                     & MSP                     & 91.0$\pm$1.4     & 79.2$\pm$2.5      & 58.8$\pm$5.3       & 94.7$\pm$0.4     & 88.1$\pm$3.0      & 74.8$\pm$7.9       \\
       \rowcolor{var3dark}          FT-T           & MLS                     & 91.2$\pm$1.0     & 80.4$\pm$3.7      & 61.0$\pm$7.7       & 94.5$\pm$0.5     & 87.9$\pm$5.8      & 79.2$\pm$13.8      \\
        \rowcolor{var3dark}                    & TempScale               & 91.0$\pm$1.4     & 79.2$\pm$2.5      & 58.8$\pm$5.3       & 94.7$\pm$0.4     & 88.0$\pm$3.0      & 74.8$\pm$7.9       \\
         \rowcolor{var3dark}                   & ODIN                    & 90.9$\pm$1.4     & 79.0$\pm$2.4      & 58.4$\pm$5.1       & 94.8$\pm$0.4     & 88.1$\pm$3.0      & 74.6$\pm$7.7       \\
         \rowcolor{var3dark}                   & EBO                     & 91.6$\pm$0.8     & 82.0$\pm$5.1      & 63.7$\pm$9.0       & 94.4$\pm$0.7     & 88.1$\pm$6.0      & 80.3$\pm$14.3      \\
         \rowcolor{var3dark}                   & GRAM                    & 94.1$\pm$0.6     & 89.8$\pm$4.7      & 79.9$\pm$8.3       & 95.1$\pm$0.3     & 88.9$\pm$2.8      & 77.9$\pm$7.0       \\
          \rowcolor{var3dark}                  & GradNorm                & 91.3$\pm$1.4     & 78.6$\pm$4.5      & 58.3$\pm$6.7       & 94.5$\pm$0.5     & 87.2$\pm$3.7      & 74.1$\pm$8.1       \\
          \rowcolor{var3dark}                  & ReAct                   & 91.8$\pm$0.7     & 82.9$\pm$3.9      & 65.5$\pm$6.7       & 94.1$\pm$0.8     & 87.2$\pm$5.3      & 78.3$\pm$11.8      \\
           \rowcolor{var3dark}                 & DICE                    & 88.3$\pm$1.1     & 66.9$\pm$2.3      & 45.1$\pm$2.7       & 92.6$\pm$1.6     & 69.5$\pm$2.4      & 48.3$\pm$3.8       \\
             \rowcolor{var3dark}               & ASH                     & 91.6$\pm$0.7     & 81.0$\pm$4.5      & 61.0$\pm$6.9       & 94.6$\pm$0.6     & 88.2$\pm$5.0      & 80.4$\pm$13.6     \\ \bottomrule
\end{tabular}
\caption{
Comparing OOD detection methods in the \textit{synthesized-OOD} setting using eICU and MIMIC-IV datasets, with performance evaluated through FPR@95. This Table is similar to Table \ref{tab:synthesized_ood}, but assesses FPR@95 instead of utilizing AUROC. 
In this setting, one feature is multiplied by a specific factor $\mathcal{F}$ to construct an OOD sample. For each model and factor we display the average performance, calculated over 100 resampling of the feature and 5 different initializations of the model. 
%Results are averaged over 5 runs.
For each model, the best value in each column is shown in bold, and distance-based post-hoc approaches are distinguished with a lighter color.
}
\label{tab:synthesized_ood_fpr}
\end{table*}

\end{document}